\RequirePackage{fix-cm}

\documentclass[aos,preprint]{imsart}

\RequirePackage{amsthm,amsmath,amsfonts,amssymb}
\RequirePackage[numbers]{natbib}

\usepackage{mathtools}
\usepackage{dsfont}
\usepackage{graphicx} 

\usepackage[algo2e,ruled,vlined,linesnumbered]{algorithm2e}
\usepackage{xspace}
\usepackage{xcolor}
\usepackage{color}

\theoremstyle{plain}
\newtheorem{theorem}{Theorem}
\newtheorem{lemma}{Lemma}[section]
\newtheorem{example}{Example}[section]
\newtheorem{proposition}{Proposition}[section]
\newtheorem{corollary}[theorem]{Corollary}

\theoremstyle{definition}
\newtheorem{definition}{Definition}[section]

\newtheorem{remark}{Remark}[section]

\newcommand{\lasso}{\textsc{Lasso}\xspace}
\newcommand{\snr}{\textsc{SNR}}
\newcommand{\nInfSp}{n_{\text{INF}}^{\text{SP}}}

\newcommand{\nInf}{n_{\text{INF}}}

\newcommand{\N}{\mathbb{N}}
\newcommand{\R}{\mathbb{R}}

\newcommand{\T}{\intercal}

\newcommand{\var}{\operatorname{Var}}
\newcommand{\cov}{\operatorname{Cov}}

\newcommand{\pr}[1]{\left({#1}\right)}

\newcommand{\prbig}[1]{\big({#1}\big)}
\newcommand{\prBig}[1]{\Big({#1}\Big)}

\newcommand{\br}[1]{\left[{#1}\right]}

\newcommand{\ac}[1]{\left\{{#1}\right\}}

\newcommand{\acn}[1]{\{{#1}\}}

\newcommand{\norm}[1]{\left\|{#1}\right\|}

\newcommand{\abs}[1]{\left\lvert{#1}\right\rvert}

\newcommand{\calC}{\ensuremath{\mathcal{C}}}

\newcommand{\calE}{\ensuremath{\mathcal{E}}}

\newcommand{\calN}{\ensuremath{\mathcal{N}}}
\newcommand{\calO}{\ensuremath{\mathcal{O}}}

\newcommand{\calS}{\ensuremath{\mathcal{S}}}

\newcommand{\inner}[2]{\langle {#1} , {#2} \rangle}
\newcommand{\symdiff}[2]{{#1} \, \triangle \, {#2}}
\newcommand{\support}[1]{\text{Supp}\pr{{#1}}}
\DeclareMathOperator*{\argmin}{argmin}

\newcommand{\E}{\mathbb{E}}

\newcommand{\Pro}{\mathbb{P}}

\newcommand{\dequal}{\stackrel{d}{=}}
\newcommand{\indic}{\mathds{1}}

\startlocaldefs

\endlocaldefs

\begin{document}

\begin{frontmatter}
\title{The Price of Sparsity: Sufficient Conditions for Sparse Recovery using Sparse and Sparsified Measurements\thanksref{T1}}
\runtitle{The Price of Sparsity}

\thankstext{T1}{A preliminary version of this work appeared in the proceedings of NeurIPS 2025~\cite{chaabouni2025price}.}

\begin{aug}
\author[A]{\fnms{Youssef}~\snm{Chaabouni}\ead[label=e1]{youss404@mit.edu}}
\and
\author[A]{\fnms{David}~\snm{Gamarnik}\ead[label=e2]{gamarnik@mit.edu}}
\address[A]{Massachusetts Institute of Technology \printead[presep={ ,\ }]{e1,e2}}
\end{aug}

\begin{abstract}
We consider the problem of support recovery for sparse binary signals from noisy linear measurements. For sparse Gaussian measurement matrices we identify sufficient conditions on the minimal sample size for maximum-likelihood recovery in the high-SNR regime \(ds/p\to\infty\), where \(p\) denotes the signal dimension, \(s\) the number of non-zero components of the signal, and \(d\) the expected number of non-zero components per row of measurement. Combined with known lower bounds, this yields an information-theoretic threshold of order \(s\log(p/s)/\log(ds/p)\), making explicit the price of measurement sparsity. In particular, we highlight a regime where the sample-complexity loss from measurement sparsity is logarithmic while the computational gain is nearly linear.

Second, we study recovery after sparsifying an originally dense Gaussian design: the observations are generated from the dense design, while estimation uses an independently sparsified design and a rescaled response. In the proportional regime \(s=\alpha p\), \(d=\psi p\), we prove that, for every fixed target error level \(\delta\) and every slack \(\varepsilon>0\), a sample size of order \(p/\psi^2\) is sufficient for support recovery for arbitrarily small \(\psi\).
\end{abstract}



\end{frontmatter}
\tableofcontents


\section{Introduction}\label{section:introduction}

In recent years, sparse signal recovery has gained significant attention, motivated by applications in compressive sensing \cite{foucart2013invitation, candes2006robust,donoho2006compressed}; signal denoising \cite{chen2001atomic}; sparse regression \cite{miller2002subset}; data stream computing \cite{cormode2009finding, indyk2007sketching, muthukrishnan2005data}; combinatorial group testing \cite{du1999combinatorial}; etc. Practical examples range from the single‑pixel camera, MRI scanners and radar remote‑sensing systems to error‑correction schemes in digital communications and widely used image‑compression formats \cite[Chap. 1]{foucart2013invitation}.

The problem can be formulated as follows. Consider a \textit{signal} \(\beta^\star \in \R^p\), unknown but a priori \(s\)-sparse for some given \(s \leq p\), a random \textit{measurement matrix} \(X \in \R^{n \times p}\) (also referred to as \textit{design}, \textit{features} or \textit{data}) and a \textit{noise} vector \(Z \sim \mathcal{N}(0, \sigma^2 I_n)\), where \(n \in \N\) denotes the \textit{sample size} and \(\sigma^2 > 0\) a fixed constant. A vector of \textit{observations} (also known as \textit{labels} or \textit{annotations}) is given by:
\[
Y \coloneqq X \beta^\star + Z.
\]
Sparse recovery refers to reconstructing \(\beta^\star\) given \(X\) and \(Y\). Intuitively, this problem can be reduced to recovering the support $S^\star$ of $\beta^\star$, i.e. the set of indices of its non-zero components. In fact, once the support $S^\star$ is identified, the full signal can be estimated using the corresponding columns of $X$ via the closed-form maximum-likelihood estimator formula $\beta_{\text{MLE}} = X_{S^\star}^+ Y$, where $X_{S^\star}^+$ denotes the Moore-Penrose pseudoinverse of the submatrix formed by the columns of $X$ with indices in $S^\star$ \cite{hastie2009elements}.

Traditionally, \(X\) was assumed to be a \textbf{dense} random matrix with sub-Gaussian entries. Previous works have shown that the complexity of the problem in terms of required sample size exhibits two phase transitions at two thresholds \(\nInf < n_{\text{ALG}}\), yielding three regimes:
\begin{itemize}
    \item \(n < \nInf\): impossibility of recovery. Reeves et al. \cite{reeves2019all} show that if \(n \leq \pr{1-\varepsilon} \nInf\) then the recovery of any fraction of the support of the signal is information-theoretically impossible.
    \item \(\nInf < n < n_{\text{ALG}}\): super-polynomial complexity. Gamarnik and Zadik \cite{gamarnik2022sparse} show that if \(n \geq \pr{1+\varepsilon} \nInf\) then the maximum-likelihood estimator (MLE) recovers the support of  \(\beta^\star\). Although solvable, the problem is widely believed to be algorithmically hard since the MLE exhibits an Overlap Gap Property (OGP) \cite{gamarnik2022sparse}.
    \item \(n > n_{\text{ALG}}\): polynomial-time recovery. Wainwright \cite{wainwright2009sharp} shows that if \(n \geq \pr{1+\varepsilon} n_{\text{ALG}}\) then the \lasso \cite{tibshirani1996regression}, which is a polynomial-time algorithm, succeeds in recovering the support of \(\beta^\star\).
\end{itemize}

\subsection{Sparse measurement setting}
While dense matrices offer an optimal sample size, they are costly in terms of storage and computation. \textbf{Sparse} measurement matrices, where the number of non-zero entries per measurement vector scales significantly smaller than the signal dimension, mitigate these costs: they require significantly less storage and allow for more efficient computations, as matrix-vector multiplications and incremental updates can be performed faster. In addition, they enable efficient signal recovery algorithms by taking advantage of the structural properties of the problem \cite{gilbert2010sparse}. However, this sparsity comes at the cost of increased sampling complexity \cite{wang2010information}. This raises the following key question: How does measurement sparsity trade off with sampling complexity?

Some of the prior studies have explored this sparse measurement setting. Wang et al. \cite{wang2010information} establish necessary conditions for sparse recovery for various measurement sparsity regimes. Let \(d\) denote the expected number of non-zero components of a row of \(X\). Their work reveals three regimes of behavior depending on \(ds/p\), the expected number of non-zero components of \(\beta^\star\) that align with non-zero components of a row of \(X\). The three regimes are: \(ds/p \rightarrow +\infty\), \(ds/p = \tau\) for some constant \(\tau > 0\), and \(ds/p \rightarrow 0\). They show that in each regime, the number of samples \(n\) must exceed a specific information-theoretic lower bound for any algorithm to reliably recover the signal's support. In particular, in the first regime, when \(ds/p \rightarrow +\infty\), the necessary condition threshold of \cite{wang2010information} is the same as the one of the dense case, while it increases dramatically in the third regime, where \(ds/p \rightarrow 0\). They work with entries rescaled so that \(\var\pr{X\beta^\star}\) matches the dense case, while we keep \(\var\pr{X_{ij}} = 1\). The settings are equivalent since any scaling of \(X\) can be accounted for in \(\beta^\star\).

In this work, we examine the opposite question: how many samples are enough to guarantee a reliable recovery? For simplicity, we assume the signal is binary, i.e. $\beta^\star \in \ac{0,1}^p$. Note that in this case, recovering the support is equivalent to recovering the signal. This assumption is very common in the literature \citep{5571873, reeves2019all, gamarnik2022sparse}. Intuitively, detecting a component of size $1$ is at least as hard as detecting a stronger component, so the resulting thresholds are representative of signals with non-zero entries bounded away from zero by $1$, i.e. \(\beta^\star \in \ac{\beta \in \R^p \colon \norm{\beta}_0 = s \text{ and } \min_{j \in [p] \,:\, \beta_j \ne 0} \abs{\beta_j} \geq 1}\). Our first main result (Theorem \ref{theorem:sparseRecoveryForSparseMeasurements}) states that in the high signal-to-noise ratio (SNR) regime where \(ds/p \rightarrow +\infty\), if the number of samples \(n\) is larger than a threshold given by:
\begin{equation}\label{eq:nInfSpIntro}
\nInfSp = \frac{2s\log\pr{p/s}}{\log\pr{ds/p}},
\end{equation}
then the MLE asymptotically recovers the support of the signal. 
The proof uses a Chernoff bound on the mean-squared-error difference between a competing support and the true one, followed by a union bound over supports. The sharpness of the threshold \eqref{eq:nInfSpIntro} follows from a tight asymptotic analysis of the row moment generating function, which exploits the conditional Gaussianity of measurement rows given the random sparsity pattern of their entries.
Bringing our result together with the necessary condition shown by Wang et al. \cite{wang2010information}, we reveal that the problem exhibits a phase transition -- similar to the one known in the dense case -- at the information-theoretic threshold \(\nInfSp\). In particular, if there exists a constant \(\varepsilon > 0\) such that \(n \leq \pr{1-\varepsilon} \nInfSp\) then it is information-theoretically impossible to ensure a reliable recovery of the support of the signal, and if there exists a constant \(\varepsilon > 0\) such that \(n \geq \pr{1+\varepsilon} \nInfSp\) then the MLE ensures a reliable recovery of the support. Our findings therefore answer the question of exactly how much data is needed for recovery. However, the two bounds refer to different notions of recovery: the necessity statement negates \emph{exact} support recovery uniformly over signals, whereas our sufficiency statement establishes only vanishing \emph{fractional} Hamming error in probability for the MLE; we discuss this gap in Remark~\ref{remark:limitationOfSparseSettingWork}.
We call the amount of additional observations in the sparse setting compared to the dense one  \emph{price of sparsity}. Precisely, restricting each measurement to \(d\) non-zeros inflates the required sample size by a factor of \(\Gamma = \log{s}/\log\pr{ds/p}\),
quantifying the sampling complexity vs. measurement sparsity trade-off.
In particular, we note that in the proportional regime $s = \Theta\pr{p}$, $d = \Theta\pr{p}$, this factor becomes negligible.

Regarding the computational complexity, Omidiran and Wainwright \cite{omidiran2008high} show that the \lasso performs as well in the sparse setting as in the dense setting, assuming a slow decay of sparsity. They show that, under some slow sparsity assumption, it is sufficient for the sample size \(n\) to be larger than the algorithmic threshold of the dense setting discussed above, given by:
\[
n_{\text{ALG}} = 2 s\log\pr{p-s},
\]
specifically for the \lasso to ensure a reliable polynomial time recovery of \(\beta^\star\). Although the sparsity assumption under which this result holds allows for the density rate \(d/p\) to go to \(0\) as \(p \rightarrow +\infty\), it still doesn't allow the measurements to be very sparse. In fact, it requires that:
\begin{equation}\label{eq:slowSparsityAssumption}
d/p = \omega\pr{s^{-1/3}}
\quad
\text{and}
\quad
d/p = \omega\pr{\pr{\frac{\log{\log\pr{p-s}}}{\log\pr{p-s}}}^{1/3}}.
\end{equation}
This raises a question about what happens in a sparser regime. Although our work does not address algorithmic questions, our sufficiency result extends to a strictly broader sparsity regime than \eqref{eq:slowSparsityAssumption}, leaving open the possibility of corresponding polynomial-time improvements in this regime.

\subsection{Active Sparsification}
The applications of the signal recovery problem \cite[Chapter 1]{foucart2013invitation} considered in this paper can be broadly categorized into two classes:
\begin{itemize}
    \item[-] Applications where \(X\) is \textbf{designed}, e.g. involving signal compression and reconstruction.
    \item[-] Applications where \(X\) is \textbf{observed}, e.g. sparse regression, signal denoising, and error correction.
\end{itemize}
In light of this categorization, we note that measuring the trade-off between measurement sparsity and sampling complexity is particularly useful for the first class of problems. It provides practitioners with an exact description of how the measurement matrix should be designed, in terms of size and sparsity, to optimize the computational cost of signal recovery. However, this is rendered useless in the second class of problems when the measurement matrix is observed and dense. This motivates the second key question: given an initially dense measurement matrix, is there a way to make it sparse and still aim to recover the original signal?

The question of whether a dense object can be substantially sparsified post-hoc without compromising a downstream task also arises in the neural network compression literature, going back to the Optimal Brain Damage (OBD) framework of LeCun, Denker and Solla~\cite{lecun1989optimal} and its second-order refinement, Optimal Brain Surgeon (OBS) of Hassibi and Stork~\cite{hassibi1992second}: there, one asks whether a trained network's weights can be largely zeroed out with little loss in predictive performance, and recent theoretical work has obtained post-training pruning guarantees for wide multilayer perceptrons~\cite{elcheairi2025theoretical}. The object being sparsified differs (estimator parameters there, the measurement matrix here) but the post-hoc sparsification question is shared.

Concretely, we model sparsification as follows. Given a dense Gaussian design \(X\), we form a sparsified design \(\Tilde{X}\) by setting each entry of \(X\) to zero independently with probability \(1-d/p\), keeping it unchanged otherwise. Equivalently, \(\Tilde{X}_{ij} \coloneqq B_{ij} X_{ij}\) where $B_{ij}, i \in [n], j \in [p]$ are i.i.d. $\text{Ber}\pr{d/p}$ random variables independent of \(X\), and \(d \leq p\) controls the sparsification rate. The dense observations \(Y = X\beta^\star + Z\) are then rescaled accordingly to form \(\Tilde{Y} \coloneqq \pr{d/p}\, Y\), and recovery is attempted from \(\pr{\Tilde{X}, \Tilde{Y}}\).

This setting, in which the observations are generated with a dense measurement matrix, but the signal is recovered using a sparsified version of it, is closely related to the ``missing covariates'' or ``missing-at-random'' framework studied in high-dimensional statistics. Prior work by Loh and Wainwright \cite{loh2011high}, established algorithmic $\ell_2$-error bounds for regression under this model, assuming dense Gaussian designs and a constant missingness rate. Our analysis in Section \ref{section:improvingSparseRecoveryViaSparsification} complements this line of research by focusing instead on information-theoretic support-recovery thresholds and deriving the precise sample complexity cost incurred by sparsification in this regime.

In our examination of the sparsification question, we focus on the linear sparsity and strong linear sparsification regime where \(s = \alpha p\) and \(d = \psi p\), with \(\alpha\in(0,1)\) fixed and \(\psi>0\) fixed and sufficiently small. Specifically, our second main result (Theorem \ref{theorem:sparsificationLinear}) states that, for every fixed error tolerance \(\delta\in(0,1)\) and every slack \(\varepsilon>0\), there exists \(\psi_0=\psi_0(\alpha,\delta,\varepsilon)>0\) such that, for every fixed \(\psi\in(0,\psi_0)\), a sample size \(n\) larger than the threshold
\begin{equation}\label{expressionOfSparsificationThresholdIntro}
\frac{2 h\pr{\alpha} p}{\log\pr{1+\frac{\delta \psi^2}{\pr{1-\psi}\pr{2-\delta\pr{1-\psi}}}}}
\end{equation}
suffices for the minimizer of the mean squared error (MSE) based on the sparsified measurements and accordingly-rescaled observations to recover the true support up to error fraction \(\delta\).
The proof of Theorem~\ref{theorem:sparsificationLinear} is substantially more involved than that of Theorem~\ref{theorem:sparseRecoveryForSparseMeasurements} because the rescaled observations $\Tilde{Y}$ are a rescaling of the original observations $Y$, not a noisy projection of the true signal through the sparsified design $\Tilde{X}$. As a consequence, the row moment generating function arising in the Chernoff bound depends on the realization of the random sparsification mask, and the most natural choice of Chernoff parameter introduces a degeneracy on an exponentially small set of masks that, treated directly, would force the analysis through an unverified uniform-integrability hypothesis. We resolve this by evaluating the Chernoff bound at a \emph{shrunken} Chernoff parameter: a regularized choice that removes the degeneracy uniformly over masks at the cost of a controlled slack in the sample-complexity bound, which the assumption $n \geq (1+\varepsilon)n^\star_{\text{SP}}$ absorbs. The smallness condition on $\psi$ is precisely the cost of this regularization, not a claim that larger $\psi$ is intrinsically harder. We believe this regularized Chernoff parameter device is of independent methodological interest.
In the strong-sparsification regime where \(\psi \rightarrow 0\), the sufficient threshold \eqref{expressionOfSparsificationThresholdIntro} effectively writes:
\begin{equation}\label{expressionOfnInfSpifiedIntro}
n_{\text{INF}}^{\text{SP}}
=
\Theta\pr{\frac{p}{\psi^2}}.
\end{equation}
We call the amount of additional observations in the sparsification setting compared to the dense one \emph{price of sparsification}. Unlike the price of sparsity, it is not due to the sparsity of the measurements but rather to a bias in the observations. We also interpret our result as providing an expression of the \emph{sparsification budget}: the level up to which one could sparsify their data and still recover the true signal. Inverting the sample-complexity bound \eqref{expressionOfSparsificationThresholdIntro}, in the regime \(n = \Omega\pr{p}\), recovery upon active sparsification holds as long as \(\psi \geq c\sqrt{p/n}\), for some constant $c > 0$. Explicitly: a practitioner with $n$ observations may zero out all but an order-$\sqrt{p/n}$ fraction of the design's entries (on average, per row) and still recover the signal. Consequently, doubling the sample size buys an additional factor of $\sqrt{2}$ in terms of sparsification budget.

\subsection{Use of large language models}

During the development of this work, the authors used large language models (Anthropic's Claude Opus 4.7) as a discussion partner for some of the technical development. In particular, the idea of evaluating the Chernoff bound at the shrunken parameter \(\theta_{p,\lambda} = \lambda\theta^\star\) for \(\lambda \in (0,1)\), which underlies Lemma~\ref{lemma:UniformFinitenessSparsification} and is the technical device that eliminates the uniform-integrability hypothesis present in the conference version, emerged from such discussions. All mathematical statements have been fully verified by the authors, who take sole responsibility for the correctness of the paper.

\subsection{Outline and Notations}

We organize the rest of the paper as follows.
Section \ref{section:sparseRecoveryUsingSparseMeasurements} studies the sparse measurement setting.
Section~\ref{section:improvingSparseRecoveryViaSparsification} examines recovery after sparsifying an originally dense measurement matrix.
Section~\ref{section:proofsForSparseSetting} contains the proofs of Theorem~\ref{theorem:sparseRecoveryForSparseMeasurements} and Corollary~\ref{corollary:informationTheoreticPhaseTransition}.
Section~\ref{section:proofOftheorem:sparsificationLinear} contains the proof of Theorem~\ref{theorem:sparsificationLinear}.
Section~\ref{section:conclusion} concludes and sketches future work directions.

Throughout this document, we will use the following notations. We denote by $h\pr{\cdot}$ the binary entropy: $h\pr{x} = - x \log{x} - \pr{1-x} \log\pr{1-x}$, $x \in \pr{0,1}$. We call \emph{$\ell_0$-norm} the number of non-zero coordinates of $x \in \R^d$, that is $\norm{x}_0 \coloneqq \sum_{i=1}^{d} \indic\pr{x_i \ne 0}$. We call \emph{support of \(u \in \R^p\)} the set of indices of the non-zero components of \(u\) and denote it \(\support{u} \coloneqq \acn{i \in [p] \colon u_i \ne 0}\), so that \(\abs{\support{u}} = \norm{u}_0\).  We call \textit{symmetric difference} between two sets \(S_1\) and \(S_2\) the set of elements in one but not the other and denote it
\(
\symdiff{S_1}{S_2} \coloneqq \prbig{S_1 \cup S_2} \setminus \prbig{S_1 \cap S_2}.
\)

\section{Sparse Recovery using Sparse Measurements}
\label{section:sparseRecoveryUsingSparseMeasurements}
\subsection{Setting}
\label{section:settingSparseMeasurements}
Let \(n, p, s, d \in \N\) such that \(d, s \leq p\). We define a sparse Gaussian matrix in \(\R^{n \times p}\) as follows.
\begin{definition}[Sparse Gaussian matrix]
\label{def:sparseGaussianMatrix}
     We call \(X = \br{X_{ij}}_{i \in [n], j \in [p]} \in \R^{n \times p}\) a sparse Gaussian matrix with parameter \(d\) if for all \(i \in [n], j \in [p]\) we have:
    \[
    X_{ij} = B_{ij} N_{ij},
    \]
    where \(\pr{B_{ij}}_{i\in[n],j\in[p]} \stackrel{\text{i.i.d.}}{\sim}  \text{Ber}\pr{d/p}\) and \(\pr{N_{ij}}_{i\in[n],j\in[p]} \stackrel{\text{i.i.d.}}{\sim} \calN\pr{0,1}\) are mutually independent, Gaussian random variables. Note that \(d\) is the expected number of non-zero components per row of \(X\). In our setting, we assume \(d\) to be of smaller order of magnitude than \(p\), i.e. \(d = o\pr{p}\).
\end{definition}

Let \(X\) be a sparse Gaussian random matrix of parameter \(d\), and \(Z\) be a random vector in \(\R^n\) such that \(Z \sim \calN\pr{0, \sigma^2 I_n}\), with \(\sigma > 0\) a fixed constant. Let \(\beta^\star \in \ac{0,1}^p\) be a deterministic vector such that \(\norm{\beta^\star}_0 = s\). We define the random vector \(Y\) as:
\begin{equation}
\label{eq:model}
    Y \coloneqq X\beta^\star + Z.
\end{equation}
Of particular interest is the \emph{signal-to-noise ratio} ($\snr$), known to be an important quantity for characterizing the difficulty of sparse recovery problems \citep{wang2010information, reeves2019all}. It's defined as follows:
\begin{equation}\label{equation:SNRGenericDefinition}
    \snr
    \coloneqq
    \frac{\E{\norm{X\beta^\star}_2^2}}{\E{\norm{Z}_2^2}}
    = \frac{ds}{p \sigma^2}.
\end{equation}
The maximum likelihood estimator (MLE) of \(\beta^\star\) is defined by the random vector:
\begin{equation}\label{eq:MLEdefinition}
\hat{\beta} \coloneqq \argmin_{\beta \in \ac{0,1}^p, \, \norm{\beta}_0 = s} \norm{Y - X\beta}_2^2.
\end{equation}
We are interested in the minimum number of samples \(n\) required so that the MLE \eqref{eq:MLEdefinition} asymptotically recovers the true signal $\beta^\star$. Specifically: given an error tolerance \(\delta \in \pr{0,1}\), we wish to determine the minimum number of samples \(n\) as a function of \(p\), \(s\) and \(d\) required so that:
\[
\Pro_{X,Z}\pr{\abs{\symdiff{\support{\beta^\star}}{\support{\hat{\beta}}}} < 2 \delta s} \longrightarrow 1,
\;\; \text{as } n,p,s,d \rightarrow +\infty.
\]

\subsection{Results}\label{section:sparseMeasurementsResults}
Our first main result, Theorem \ref{theorem:sparseRecoveryForSparseMeasurements}, provides a sufficient condition on the sample size for reliable support recovery when using sparse measurements.

\begin{theorem}[Sufficient conditions for sparse recovery using sparse measurement matrices]
\label{theorem:sparseRecoveryForSparseMeasurements}
    Suppose \(p,s,d \rightarrow +\infty\), \(d = o\pr{p}\) and \(ds = \omega\pr{p}\) (i.e. $\snr \rightarrow +\infty$). Let \(\delta \in \pr{0,1}\).
    We consider two different regimes.
    \begin{enumerate}
        \item Assume \(s = o\pr{p}\). Let
        \[
        n^\star_{\text{SP}} \coloneqq \frac{2 s \log\pr{p/s}}{\log\pr{ds/p} + \log\pr{\delta/(2\sigma^2)}}.
        \]
        If there exists \(\varepsilon > 0\) such that \(n \geq \pr{1+\varepsilon} n^\star_{\text{SP}}\), then the MLE \(\hat{\beta}\) recovers \(\beta^\star\) up to error \(\delta\) w.h.p.:
        \[
        \Pro_{X,Z}\pr{\abs{\symdiff{\support{\beta^\star}}{\support{\hat{\beta}}}} < 2 \delta s}
        \geq
        1 - \exp\Big(- \varepsilon s \log\pr{p/s} + o\,\big(s \log\pr{p/s}\big)\Big),
        \]
        as \(n,p,s,d \rightarrow +\infty\).
        \item Assume there exists a constant \(\alpha \in \pr{0,1}\) such that \(s = \alpha p\). Let:
        \[
        n^\star_{\text{SP}} \coloneqq \frac{2 h\pr{\alpha} p}{\log{d} + \log\pr{\delta \alpha/(2\sigma^2)}},
        \]
        where \(h\pr{\cdot}\) denote the entropy function. If there exists \(\varepsilon > 0\) such that \(n \geq \pr{1+\varepsilon} n^\star_{\text{SP}}\), then the MLE \(\hat{\beta}\) recovers \(\beta^\star\) up to error \(\delta\) w.h.p.:
        \[
        \Pro_{X,Z}\pr{\abs{\symdiff{\support{\beta^\star}}{\support{\hat{\beta}}}} < 2 \delta s}
        \geq
        1 - \exp\Big(- \varepsilon h\pr{\alpha} p + o\pr{p}\Big),
        \]
        as \(n,p,s,d \rightarrow +\infty\).
    \end{enumerate}
\end{theorem}

The proof of Theorem \ref{theorem:sparseRecoveryForSparseMeasurements}, given in section \ref{section:proofOftheorem:sparseRecoveryForSparseMeasurements}, uses large deviation techniques to bound the probability of a high-error support to have a lower MSE than the true one, then a union bound over such supports. We give below a brief proof sketch of Theorem \ref{theorem:sparseRecoveryForSparseMeasurements}.
\begin{proof}[Proof sketch]
Let \(\calS\) denote the set of supports of cardinality \(s\) and \(S^\star = \support{\beta^\star}\). For any \(S \in \calS\), we denote by \(\indic_S\) the vector in \(\ac{0,1}^p\) such that \(\br{\indic_S}_i = \indic\pr{i \in S}\) for all \(i \in [p]\). We define the loss function \(L\) over \(\calS\) such that \(L\pr{S} \coloneqq \norm{Y-X\indic_S}_2^2\), so that \(\support{\hat{\beta}} = \argmin_{S \in \calS} L\pr{S}\). As \(p\) gets large, the event ``\(L\pr{S} < L\pr{S^\star}\)'' for any \(S\) such that \(\abs{\symdiff{S}{S^\star}} \geq 2 \delta s\) is a \emph{rare event}. The Chernoff bound yields:
\[
\log \Pro\prBig{L\pr{S} < L\pr{S^\star}} \leq \frac{n}{2} \pr{\log\pr{\frac{2 \sigma^2 p}{\delta ds}} + o\pr{1}}.
\]
This step involves most of the technical work. Then, by union bound:
\begin{align*}
\Pro\pr{\abs{\symdiff{\support{\hat{\beta}}}{S^\star}} < 2 \delta s}
&\geq 1 - \sum_{S\colon \abs{\symdiff{S}{S^\star}} \geq 2 \delta s} \Pro\pr{L\pr{S} < L\pr{S^\star}}\\
&\geq 1 - \binom{p}{s} \pr{\frac{2\sigma^2 p}{\delta d s}}^{n/2}.
\end{align*}
Solving for \(n\), we obtain a critical threshold of \(n^\star = \frac{2 \log\binom{p}{s}}{\log\pr{ds/p} + \log\pr{\delta/\pr{2\sigma^2}}}\). We conclude.
\end{proof}

Bringing together Theorem \ref{theorem:sparseRecoveryForSparseMeasurements} with the necessary conditions shown by Wang et al. in \cite{wang2010information}, we obtain the following corollary.
\begin{corollary}[Information-theoretic phase transition]
\label{corollary:informationTheoreticPhaseTransition}
The sparse recovery in the sparse setting problem exhibits a phase transition at an information-theoretic threshold \(\nInfSp\). 
\begin{enumerate}
    \item In the first regime considered above, the expression of \(\nInfSp\) is given by:
    \[
    \nInfSp \coloneqq \frac{
    2 s \log\pr{p/s}}{\log\pr{ds/p}}.
    \]
    \item In the second regime considered above, the expression of \(\nInfSp\) is given by:
    \[
    \nInfSp \coloneqq \frac{
    2 h\pr{\alpha} p}{\log{d}}.
    \]
\end{enumerate}
Specifically, in each of these regimes:
\begin{itemize}
    \item[(\textit{i})]
    If there exists \(\varepsilon > 0\) such that \(n \leq \pr{1-\varepsilon} \nInfSp\) then, as \(n,p,s,d \rightarrow +\infty\),  there exists no decoder \(g \colon \R^n \rightarrow \ac{\beta \in \ac{0,1}^{p} \colon \norm{\beta}_0 = s}\) such that:
    \[
    \max_{\beta^\star \in \ac{0,1}^p,\, \norm{\beta^\star}_0=s}
    \Pro_{X,Z}\prBig{g\pr{Y} \ne \support{\beta^\star}}
    \rightarrow 0.
    \]
    In this sense, it is information-theoretically impossible to ensure an asymptotically reliable recovery. 
    
    \item[(\textit{ii})] If there exists \(\varepsilon > 0\) such that \(n \geq \pr{1+\varepsilon} \nInfSp\), then as \(n,p,s,d \rightarrow +\infty\):
    \[
    \frac{\abs{\symdiff{\support{\hat{\beta}}}{\support{\beta^\star}}}}{2s}
    \longrightarrow 0,
    \]
    in probability. In this sense, the MLE \eqref{eq:MLEdefinition} ensures an asymptotically reliable recovery.
\end{itemize}
\end{corollary}

The proof of Corollary \ref{corollary:informationTheoreticPhaseTransition} is given in section \ref{section:proofOfcorollary:informationTheoreticPhaseTransition}. Statement (\textit{i}) is due to Wang et al. \cite{wang2010information}, while statement (\textit{ii}) follows from Theorem \ref{theorem:sparseRecoveryForSparseMeasurements} and is the main contribution of this section.

\begin{remark}[Limitations]
\label{remark:limitationOfSparseSettingWork}
Statements (\textit{i}) and (\textit{ii}) refer to different notions of recovery: (\textit{i}), due to~\cite{wang2010information}, negates \emph{exact} support recovery uniformly over signals, while (\textit{ii}) only establishes vanishing \emph{fractional} Hamming error in probability for the MLE. The two are logically compatible, so Corollary~\ref{corollary:informationTheoreticPhaseTransition} establishes a phase transition weaker than the All-or-Nothing phenomenon of Reeves, Xu and Zadik~\cite{reeves2019all}, who in the dense setting state both bounds in terms of the same quantity. We believe an analogous All-or-Nothing strengthening should hold in our setting but leave it to future work.
\end{remark}

We interpret Theorem \ref{theorem:sparseRecoveryForSparseMeasurements} and Corollary \ref{corollary:informationTheoreticPhaseTransition} as follows.
\begin{itemize}
    \item \textbf{Phase transition.} For simplicity, we only discuss the sublinear sparsity regime, defined by \(s = o\pr{p}\). Previous works on sparse recovery in the dense case (\cite{reeves2019all},\cite{gamarnik2022sparse}) have shown the existence of an information-theoretic threshold:
    \begin{equation}\label{eq:nInfDenseInfoTheoreticThres}
    \nInf = \frac{2 s \log\pr{p/s}}{\log{s}},
    \end{equation}
    at which the complexity of support recovery in terms of sample size exhibits a phase transition, where the recovery of any fraction of the support is impossible for \(n \leq \pr{1-\varepsilon} \nInf\), and full recovery is guaranteed by the MLE for \(n \geq \pr{1+\varepsilon} \nInf\). In light of this, we ask if the support recovery problem for the class of sparse measurement matrices described above exhibits a similar behavior. In Corollary \ref{corollary:informationTheoreticPhaseTransition}, we show that indeed, it exhibits a similar phase transition at an information-theoretic threshold given by:
    \begin{equation}\label{eq:nInfSpSparseInfoTheoreticThres}
    \nInfSp = \frac{2 s \log\pr{p/s}}{\log\pr{ds/p}}.
    \end{equation}
    In Table \ref{tab:complexity_summary_sublinear}, we summarize these information-theoretic thresholds alongside known algorithmic thresholds for the sublinear sparsity regime $\prbig{s=o\pr{p}}$, highlighting the comparison between dense and sparse measurements in the high-SNR setting.
    \begin{table}
    \caption{Comparison of Sample Complexity Thresholds for Sublinear Sparsity ($s=o(p)$).}
    \label{tab:complexity_summary_sublinear}
    \begin{tabular}{@{}lcccc@{}}
    \hline
    & \multicolumn{2}{c}{Info-Theoretic} & \multicolumn{2}{c}{Algorithmic} \\
    \cline{2-3} \cline{4-5}
    Measurement & Necessary & Sufficient & Necessary & Sufficient \\
    \hline
    Dense & 
    $\frac{2 s\log(p/s)}{\log s}$ \cite{reeves2019all, wang2010information} & 
    $\frac{2 s\log(p/s)}{\log s}$ \cite{gamarnik2022sparse, reeves2019all} & 
    $2s\log(p-s)$ \cite{gamarnik2022sparse} & 
    $2s\log(p-s)$ \cite{wainwright2009sharp} \\[6pt]
    
    \begin{tabular}{@{}l@{}}Sparse, \\ high SNR\end{tabular} &
    $\frac{2 s\log(p/s)}{\log(ds/p)}$ \cite{wang2010information} & 
    $\frac{2 s\log(p/s)}{\log(ds/p)}$ (Thm~\ref{theorem:sparseRecoveryForSparseMeasurements}) & 
    Unknown & 
    $2s\log(p-s)$ \cite{omidiran2008high}$^\dagger$ \\
    \hline
    \end{tabular}
    \par \vspace{2mm}
    \footnotesize{$^\dagger$Holds under the slow decay of sparsity assumption in \cite{omidiran2008high}, see \eqref{eq:slowSparsityAssumption}.}
    \end{table}

    \item \textbf{Price of Sparsity.} In particular, we notice that \(\nInfSp \geq \nInf\). This confirms the intuition that sparse recovery requires more samples in the sparse measurement case. Corollary \ref{corollary:informationTheoreticPhaseTransition} is to be interpreted as providing an exact value for the \emph{price of sparsity}, i.e. the extra amount of observations required in the sparse setting compared to the dense one, which is given by:
    \begin{equation}\label{eq:PriceOfSparsity}
    \Gamma \coloneqq \frac{\nInfSp}{\nInf} = \frac{\log{s}}{\log\pr{ds/p}} > 1.
    \end{equation}

    \item
    Note that the expression of the price of sparsity heavily depends on the regimes of \(d\) and \(s\). The smaller the density rate \(d/p\), the more ``expensive'' the desired sparsity of the measurements is, as suggested by the expression of \(\Gamma\). In particular, \(\Gamma\) could take any value in \(\pr{1,+\infty}\), depending on the regimes of \(d\) and \(s\) w.r.t. \(p\).
    \begin{example}\label{example:sparseMeasurementsPriceOfSparsityIn(1,+infty)}\rm
    Consider the setting where \(s=p^\alpha\), \(d=p^\beta\) with \(\alpha, \beta \in \pr{0,1}\) such that \(\alpha+\beta > 1\). Then \(\Gamma =
    \alpha/\pr{\alpha + \beta - 1} \in \pr{1, +\infty},
    \)
    which approaches \(1\) when \(\beta\) approaches \(1\) (low measurement sparsity), and approaches \(+\infty\) when \(\alpha\) is fixed and \(\beta\) approaches \(1-\alpha\) (high measurement sparsity).
    \end{example}

    \item Thus, we see a measurement sparsity vs. sampling complexity trade-off, which can also be interpreted as a trade-off between sampling complexity and computational cost. We consider an example that highlights this trade-off.
    \begin{example}\label{example:measurementSparsitySamplingComplexityTradeoff}\rm
    Let \(\varphi(x) \coloneqq x/\log x\) for \(x \geq e\). Consider two measurement matrices: \(X_1\) a dense Gaussian in \(\R^{n_1 \times p}\) and \(X_2\) a sparse Gaussian \(\R^{n_2 \times p}\) with only \(d = \min\pr{p^{o\pr{1}}, \varphi^{-1}\pr{o\pr{\varphi\pr{p}}}}\) expected non-zero entries per row; and an \(s\)-sparse signal \(\beta^\star \in \R^p\), in the linear sparsity regime where \(s = \alpha p\) for constant \(\alpha \in \pr{0,1}\). On the one hand, the number of samples required for reliable recovery raises from $n_1 = \nInf = \Theta\prbig{p/\log{p}}$ in the dense case to $n_2 = \nInfSp = \Theta\prbig{p/\log{d}}$ in the sparse one. On the other hand the computational cost of recovery the support is smaller in the sparse case, as matrix-vector multiplication cost drops from $n_1 p = \Theta\prbig{p^2/\log{p}}$ to $n_2 d = \Theta\prbig{pd/\log{d}}$. This highlights a trade-off between sampling complexity and computational cost. Namely, the sample-complexity ratio \(n_2/n_1\) is at most logarithmic in $p$, while the computational gain \(n_1 p / (n_2 d)\) is nearly linear in $p$. Proofs of these statements are given in section \ref{section:proofOfExample:measurementSparsitySamplingComplexityTradeoff}.
    \end{example}

    \item \textbf{Allowing for more sparsity.} Note that the sparsity assumption under which Theorem \ref{theorem:sparseRecoveryForSparseMeasurements} guarantees reliable recovery is weaker than the sparsity assumption of the sufficient algorithmic threshold of Omidiran and Wainwright \cite{omidiran2008high} which guarantees polynomial-time recovery if there exists \(\varepsilon > 0\) such that \(n \geq \pr{1+\varepsilon} n_{\text{ALG}}\). As they show, this holds under the assumption that:
    \[
    \pr{\frac{d}{p}}^{3} \min\ac{s, \frac{\log{\log\pr{p-s}}}{\log\pr{p-s}}} \rightarrow +\infty.
    \]
    For example, when \(s = \Theta\pr{p}\), this requires that \(d = \omega\pr{p^{2/3}}\), while our result holds under the weaker assumption of \(d = \omega\pr{1}\). Equivalently, our information-theoretic guarantee tolerates measurements whose number of non-zero entries per row grows arbitrarily slowly with $p$, whereas \cite{omidiran2008high} requires this number to grow polynomially. This broader sparsity regime comes at the cost of potential super-polynomial computational complexity, since computing the MLE \eqref{eq:MLEdefinition} is exponential-time in general.
\end{itemize}

\section{Sparse Recovery via Active Sparsification}
\label{section:improvingSparseRecoveryViaSparsification}

\subsection{Setting}
Let \(n, p, s, d \in \N\) and \(\beta^\star \in \ac{0,1}^p\) \(s\)-sparse, defined as in Section \ref{section:settingSparseMeasurements}. Let \(X \in \R^{n \times p}\) such that \(\pr{X_{i,j}}_{i\in[n],j\in[p]} \stackrel{\text{i.i.d.}}{\sim} \calN\pr{0,1}\) and \(Z \sim \calN\pr{0, \sigma^2 I_n}\), with \(\sigma > 0\) constant. Let \(Y \coloneqq X \beta^\star + Z \in \R^n\).
Let \(\pr{B_{ij}}_{i\in [n], j \in [p]} \stackrel{\text{i.i.d.}}{\sim} \text{Ber}\pr{d/p}\). We define the following \textit{sparsified} version of \(X\):
\begin{equation}\label{eq:sparsificationOfGaussianMatrix}
\Tilde{X} \in \R^{n \times p}
\;\;\text{such that}\;\;
\Tilde{X}_{ij} \coloneqq B_{ij} X_{ij}, \; \forall \, i \in [n]
, j \in [p].
\end{equation}
In addition, we define a rescaled version of \(Y\) as follows:
\begin{equation} \label{eq:rescalingObservations}
\Tilde{Y} \coloneqq \frac{d}{p} \,  Y \, \in \R^n.
\end{equation}
An estimator of \(\beta^\star\) is defined by the random vector:
\begin{equation}
\label{eq:MLESparsification}
\hat{\beta} \coloneqq \argmin_{\beta \in \ac{0,1}^p, \, \norm{\beta}_0 = s} \norm{\Tilde{Y} - \Tilde{X}\beta}_2^2.
\end{equation}
That is, the observations are generated with a dense measurement matrix ($X$), but the signal is recovered using a sparsification of that matrix ($\Tilde{X}$). We formalize the problem we address below as follows: given an error tolerance \(\delta \in \pr{0,1}\), we wish to determine the minimum number of samples \(n\) in terms of \(p\), \(s\) and \(d\) required so that:
\[
\Pro_{X,Z}\pr{\abs{\symdiff{\support{\beta^\star}}{\support{\hat{\beta}}}} < 2 \delta s} \longrightarrow 1,
\;\; \text{as }
n,p,s,d \rightarrow +\infty.
\]

\subsection{Results}

Our second main result, Theorem \ref{theorem:sparsificationLinear}, provides a sufficient condition on the sample size for reliable support recovery after sparsifying an originally dense measurement matrix.

\begin{theorem}[Sufficient conditions for sparse recovery using sparsified measurements]
\label{theorem:sparsificationLinear}
Fix \(\alpha,\delta\in(0,1)\). For every \(\varepsilon>0\), there exists
\(\psi_0=\psi_0(\alpha,\delta,\varepsilon)>0\) such that the following holds. Let
\(\psi\in(0,\psi_0)\) be fixed, and suppose that, as \(p\to\infty\),
\(s=\alpha p\) and \(d=\psi p\), up to harmless integer rounding. Let
\begin{equation}\label{eq:sparsificationThreshold}
n^\star_{\text{SP}}
\coloneqq
\frac{2 h\pr{\alpha} p}{\log\pr{1 + \frac{\delta \psi^2}{\pr{1-\psi}\pr{2-\delta\pr{1-\psi}}}}}.
\end{equation}
If \(n \geq \pr{1+\varepsilon} n^\star_{\text{SP}}\), then \(\hat{\beta}\) recovers \(\beta^\star\) up to error \(\delta\) w.h.p.:
\[
\Pro\pr{\abs{\symdiff{\support{\beta^\star}}{\support{\hat{\beta}}}} < 2 \delta s}
\longrightarrow 1.
\]
\end{theorem}


The proof of Theorem \ref{theorem:sparsificationLinear} is given in Section~\ref{section:proofOftheorem:sparsificationLinear}. It follows the same Chernoff-plus-union-bound outline as Theorem~\ref{theorem:sparseRecoveryForSparseMeasurements}, with the substantial difference that the row moment generating function now depends on the realization of the sparsification mask; the technical core is a \emph{regularized Chernoff parameter} device that sidesteps a vanishing-discriminant defect, sketched in detail at the start of Section~\ref{section:proofOftheorem:sparsificationLinear}.\\

We interpret Theorem \ref{theorem:sparsificationLinear} as follows.

\begin{itemize}
\item \textbf{Arbitrary sparsification rate.} According to Theorem \ref{theorem:sparsificationLinear}, for every fixed target accuracy and slack, support recovery is possible for all sufficiently small fixed \(\psi\), provided a large enough sample size.
\item \textbf{Strong-sparsification regime.} In the strong-sparsification regime where \(\psi \rightarrow 0\), the denominator of \(n^\star_{\text{SP}}\) in \eqref{eq:sparsificationThreshold} is effectively \(\delta \psi^2 / \pr{2-\delta}\), and hence the sufficient condition upper bound writes:
\begin{equation}\label{eq:sparsificationThresHighSparsification}
n^{\text{SP}}_{\text{INF}}
= \frac{2 \pr{2-\delta} h\pr{\alpha} p}{\delta\psi^2}
= \Theta\pr{\frac{p}{\psi^2}}.
\end{equation}
\item \textbf{Price of Sparsification.} We interpret our result as providing a value for the \emph{price of sparsification}, i.e. the extra amount of observations required due to the information loss resulting from sparsification. In the linear sparsity and strong-sparsification regime, it writes:
\[
\Gamma_{\text{Sparsification}}
\coloneqq
\frac{n^{\text{SP}}_{\text{INF}}}{\nInf}
= \frac{\Theta\pr{p/\psi^2}}{\Theta\pr{p/\log{p}}}
= \Theta\pr{\frac{\log{p}}{\psi^2}}.
\]
Unlike the intrinsically-sparse-observations setting studied in Section \ref{section:sparseRecoveryUsingSparseMeasurements}, this extra amount of required observations is \emph{not} due to the sparsity of the measurements. In fact, one can check from \eqref{eq:PriceOfSparsity} that in the proportional sparsity regime where \(d = \Theta\pr{p}\), the price of sparsity is negligible, i.e. \(\Gamma \rightarrow 1\).
Instead, the price of sparsification is due to a \emph{bias in the observations} that we explain by the fact that the sparsified observations \(\Tilde{Y}\) were not obtained as noisy projections of the true signal as in the original model \eqref{eq:model}, but rather via a naïve rescaling of the original observations \eqref{eq:rescalingObservations}. By simply rescaling the observations we did not discard the information in \(Y\) coming from the nullified components of \(X\), hence introducing a bias.
\item \textbf{Sparsification budget.}
Given dense data and a fixed large enough sample size \(n\), by up to how much could we sparsify the data and still get recovery? We call this the \emph{sparsification budget}. According to our result  \eqref{eq:sparsificationThresHighSparsification}, its expression in the strong-sparsification regime is given by:
\[
\psi_{\text{budget}} = \Theta\pr{\sqrt{p/n}}.
\]
In particular, the above expression only makes sense when \(n = \Omega\pr{p}\), below which Theorem \ref{theorem:sparsificationLinear} does not hold.
\end{itemize}

\section{Sparse Recovery using Sparse Measurements: Proofs}
 \label{section:proofsForSparseSetting}
\subsection{Proof of Theorem \ref{theorem:sparseRecoveryForSparseMeasurements}}
\label{section:proofOftheorem:sparseRecoveryForSparseMeasurements}
\begin{proof}[Proof of Theorem \ref{theorem:sparseRecoveryForSparseMeasurements}]
For any \(i \in [n]\), we denote by \(X_i \coloneqq \pr{X_{ij}}_{j \in [p]}\), \(B_i \coloneqq \pr{B_{ij}}_{j \in [p]}\), \(N_i \coloneqq \pr{N_{ij}}_{j \in [p]}\). We denote by \(S^\star \coloneqq \support{\beta^\star}\) the support of \(\beta^\star\).
Let \(\calS \coloneq \left\{ S \subset [p] \; \colon \; \abs{S}=s \right\}\). We define the function:
\begin{align*}
    L \; \colon \;
    &\mathcal{S} \longrightarrow [0,+\infty)\\
    &S \longmapsto \norm{Y - X \indic_S}_2^2,
\end{align*}
where \(\indic_S\) denotes the vector in \(\ac{0,1}^p\) such that \(\br{\indic_S}_j = \indic\pr{j \in S}\) for all \(j \in [p]\).
Note that, since \(X\) and \(Y\) are random, \(L(S)\) is a random variable for every \(S \in \mathcal{S}\). In addition, note that:
\[
L(S) = \norm{Z}_2^2 + \norm{X\pr{\indic_{S^\star} - \indic_S}}_2^2 + 2 \inner{Z}{X\pr{\indic_{S^\star} - \indic_S}} \quad \forall \; S \in \mathcal{S},
\]
and, in particular:
\[
L(S^\star) = \norm{Z}_2^2 = \sum_{i = 1}^{n} Z_i^2.
\]
Fix \(S \in \calS\) such that \(M \coloneqq \abs{\symdiff{S}{S^\star}} / 2 \geq \delta s\), and let \(U \coloneqq S^\star \setminus S\), \(V \coloneqq S \setminus S^\star\). Note that \(|U| = |V| = M\). We define:
\[
\Delta
\coloneqq
L(S) - L(S^\star).
\]

\begin{proposition}\label{proposition:LargeDeviationBound}
As \(n,p,s,d \rightarrow +\infty\):
\[
\Pro\pr{\Delta \leq 0} \leq \pr{\frac{2\sigma^2p}{\delta ds}}^{n/2} e^{o(n)}.
\]
\end{proposition}
\textit{Proof.} See section \ref{section:proofLemmaLargeDeviationBound}.

Hence, we obtain:
\begin{equation}\label{eq:probErrorForAnyS}
\Pro\pr{\norm{Y - X \indic_{S}}_2^2 \leq \norm{Y - X \indic_{S^\star}}_2^2}
\leq
\pr{\frac{2\sigma^2p}{\delta ds}}^{n/2} e^{o(n)},
\end{equation}
for any \(S \in \ac{0,1}^{p}\) such that \(|S| = s\) and \(\abs{\symdiff{S}{S^\star}} \geq 2 \delta s\).

Using \eqref{eq:probErrorForAnyS} and the union bound over the set of supports \(S \text{ s.t. } \abs{\symdiff{S}{S^\star}} \geq 2 \delta s\), we obtain:
\begin{align*}
&\Pro_{X, Z}\pr{\abs{\symdiff{\support{\beta^\star}}{\support{\hat{\beta}}}} < 2 \delta s}\\
&\geq
\Pro_{X, Z}\pr{\norm{Y - X \indic_{S}}_2^2 > \norm{Y - X \indic_{S^\star}}_2^2, \forall \, S \colon \abs{\symdiff{S}{S^\star}} \geq 2 \delta s}\\
&=
1 - \Pro_{X, Z}\pr{
\exists \, S \colon \abs{\symdiff{S}{S^\star}} \geq 2 \delta s, \,
\norm{Y - X \indic_{S}}_2^2 \leq \norm{Y - X \indic_{S^\star}}_2^2}\\
&\stackrel{\text{U.B.}}{\geq}
1 -
\sum_{S \, : \, \abs{\symdiff{S}{S^\star}} \geq 2 \delta s} \Pro_{X, Z}\pr{\norm{Y - X \indic_{S}}_2^2 \leq \norm{Y - X \indic_{S^\star}}_2^2}\\
&\geq
1 - \binom{p}{s} \pr{\frac{2\sigma^2p}{\delta ds}}^{n/2} e^{o(n)}.
\end{align*}

\subsubsection{Sublinear regime: \(s = o\pr{p}\)} Using the Corollary of Stirling:
\[
\log{\binom{p}{s}}
= s \log\pr{p/s} \pr{1 + o\pr{1}},
\]
in the RHS of the inequality above, we obtain:
\begin{align*}
&\Pro_{X, Z}\pr{\abs{\symdiff{\support{\beta^\star}}{\support{\hat{\beta}}}} < 2 \delta s}\\
&\geq
1 - \exp\br{s \log\pr{p/s} \pr{1+o\pr{1}} - \frac{n}{2} \pr{ \log\pr{\frac{\delta ds}{2 \sigma^2p}} + o\pr{1}}}.
\end{align*}

Let \(n^\star_{\text{SP}} \coloneqq \frac{2 s \log\pr{p/s}}{\log\pr{ds/p} + \log\pr{\delta/(2\sigma^2)}}\). Then if \(n \geq \pr{1+\varepsilon} n^\star_{\text{SP}}\) for some constant \(\varepsilon > 0\), we have:
\begin{align*}
&\Pro_{X, Z}\pr{\abs{\symdiff{\support{\beta^\star}}{\support{\hat{\beta}}}} < 2 \delta s}\\
&\geq
1 - \exp\br{s \log\pr{p/s} \pr{1+o\pr{1}} - \frac{\pr{1+\varepsilon} n^\star_{\text{SP}}}{2} \pr{ \log\pr{\frac{\delta ds}{2 \sigma^2p}} + o\pr{1}}}\\
&=
1 - \exp\br{
s \log\pr{p/s}
\pr{
- \varepsilon + o\pr{1}
- \frac{1+\varepsilon}{\log\pr{ds/p} + \log\pr{\delta/(2\sigma^2)}} \, o\pr{1}
}}\\
&=
1 - \exp\Big(
s \log\pr{p/s}
\big(
- \varepsilon + o\pr{1}
\big)
\Big).
\end{align*}
Hence, as \(n,p,s,d \rightarrow +\infty\):
\[
\Pro_{X, Z}\pr{\abs{\symdiff{\support{\beta^\star}}{\support{\hat{\beta}}}} < 2 \delta s}
\geq
1 - \exp\Big(
- \varepsilon s \log\pr{p/s}
+ o\,\big(s \log\pr{p/s}\big)
\Big) \longrightarrow 1.
\]

\subsubsection{Linear regime: \(s = \alpha p\) , \(\alpha \in \pr{0,1}\)}
Using the Corollary of Stirling:
\[
\log\binom{p}{s} = p h\pr{\alpha} \pr{1+o\pr{1}},
\]
we get:
\begin{align*}
&\Pro_{X, Z}\pr{\abs{\symdiff{\support{\beta^\star}}{\support{\hat{\beta}}}} < 2 \delta s}\\
&\geq
1 - \exp\br{p h\pr{\alpha} \pr{1+o\pr{1}} - \frac{n}{2} \pr{ \log\pr{\frac{\delta ds}{2 \sigma^2p}} + o\pr{1}}}.
\end{align*}
Similarly to above, we take \(n^\star_{\text{SP}} \coloneqq \frac{2h\pr{\alpha}p}{\log{d}+\log\pr{\delta \alpha/\pr{2\sigma^2}}}\). If \(n \geq \pr{1+\varepsilon} n^\star_{\text{SP}}\) for some constant \(\varepsilon > 0\), then we obtain, as \(p,s,d \rightarrow +\infty\):
\[
\Pro_{X, Z}\pr{\abs{\symdiff{\support{\beta^\star}}{\support{\hat{\beta}}}} < 2 \delta s}
\geq
1 - \exp\Big(
- \varepsilon 
h\pr{\alpha} p + o\pr{p}
\Big) \longrightarrow 1,
\]
concluding the proof.
\end{proof}

\subsubsection{Proof of Proposition \ref{proposition:LargeDeviationBound}}
\label{section:proofLemmaLargeDeviationBound}
\begin{proof}[Proof of Proposition \ref{proposition:LargeDeviationBound}]
We have:
\begin{align*}
\Delta
&\coloneqq
L(S) - L(S^\star)\\
&=
\norm{X\pr{\indic_{S^\star} - \indic_S}}_2^2 + 2 \inner{Z}{X\pr{\indic_{S^\star} - \indic_S}}\\
&=
\sum_{i=1}^{n} \inner{X_i}{\indic_{S^\star} - \indic_S}^2 + 2 \sum_{i=1}^{n} Z_i \inner{X_i}{\indic_{S^\star} - \indic_S}.
\end{align*}
We denote by \(\pr{\Delta_i}_{i \in [n]}\) the terms of the sum in the above expression, that is:
\[
\Delta_i \coloneqq \inner{X_i}{\indic_{S^\star} - \indic_S}^2 + 2 Z_i \inner{X_i}{\indic_{S^\star} - \indic_S}.
\]
Note that \((\Delta_i)_{i \in [n]}\) are i.i.d. and \(\Delta = \sum_{i=1}^{n} \Delta_i\). 

Now using the Chernoff bound:
\begin{equation}\label{eq:chernoff}
\Pro\pr{\Delta \leq 0}
=
\Pro\pr{ - \Delta \geq 0}
=
\inf_{\theta \geq 0} \Pro\pr{e^{-\theta\Delta} \geq 1}
\leq
\inf_{\theta \geq 0} M_{-\Delta_i}(\theta)^n.
\end{equation}
We now study the moment generating function of \(-\Delta_i\), i.e. \(M_{-\Delta_i}(\cdot)\). We have:
\begin{align*}
M_{-\Delta_i}(\theta)
&=
\E_{X_i, Z_i}\br{e^{-\theta \br{
\inner{X_i}{\indic_{S^\star}-\indic_{S}}^2
+ 2 Z_i \inner{X_i}{\indic_{S^\star}-\indic_{S}}
}}}\\
&=
\E_{X_i}\br{e^{- \theta \inner{X_i}{\indic_{S^\star}-\indic_{S}}^2} \E_{Z_i}\br{e^{-2\theta Z_i \inner{X_i}{\indic_{S^\star}-\indic_{S}}} \big| X_i}}\\
&=
\E_{X_i}\br{e^{- \theta \inner{X_i}{\indic_{S^\star}-\indic_{S}}^2} M_{Z_i | X_i}\pr{-2\theta \inner{X_i}{\indic_{S^\star}-\indic_{S}}}}\\
&=
\E_{X_i}\br{e^{- \theta \inner{X_i}{\indic_{S^\star}-\indic_{S}}^2} e^{\frac{1}{2} \pr{-2 \theta \inner{X_i}{\indic_{S^\star}-\indic_{S}}}^2 \sigma^2}}\\
&=
\E_{X_i}\br{e^{\pr{- \theta + 2 \theta^2 \sigma^2}\inner{X_i}{\indic_{S^\star}-\indic_{S}}^2}}\\
&=
\E_{X_i}\br{e^{\pr{- \theta + 2 \theta^2 \sigma^2} \pr{\sum_{j \in U} X_{ij} - \sum_{j \in V} X_{ij}}^2}}.
\end{align*}
Plugging this expression into \eqref{eq:chernoff}, we obtain:
\[
\log{\Pro\pr{\Delta \leq 0}}
\leq
n \inf_{\theta \geq 0} \log{\E_{X_i}\br{e^{\pr{- \theta + 2 \theta^2 \sigma^2} \pr{\sum_{j \in U} X_{ij} - \sum_{j \in V} X_{ij}}^2}}}.
\]
Studying the function \(\theta \mapsto -\theta + 2 \theta^2 \sigma^2\) on \(\R_{\geq 0}\) leads to the change of variable:
\begin{align}
&\inf_{\theta \geq 0} \log{\E_{X_i}\br{e^{\pr{- \theta + 2 \theta^2 \sigma^2} \pr{\sum_{j \in U} X_{ij} - \sum_{j \in V} X_{ij}}^2}}}\\
&\label{eq:changeOfVariableTheta}
=
\inf_{\theta \in (-\infty, 1/\pr{8\sigma^2}]} \log{\E_{X_i}\br{e^{-\theta \pr{\sum_{j \in U} X_{ij} - \sum_{j \in V} X_{ij}}^2}}},
\end{align}
One can check that the function \(\theta \mapsto \log{\E_{X_i}\br{e^{-\theta \pr{\sum_{j \in U} X_{ij} - \sum_{j \in V} X_{ij}}^2}}}\) is non-increasing over \((-\infty, 1/\pr{8\sigma^2}]\). Hence \eqref{eq:changeOfVariableTheta} is equal to:
\[
\log{\E_{X_i}\br{e^{-\pr{\sum_{j \in U} X_{ij} - \sum_{j \in V} X_{ij}}^2 / \pr{8\sigma^2}}}}.
\]
Therefore, the Chernoff bound yields:
\begin{equation}\label{eq:logErrLeqNLogMgf}
\log{\Pro\pr{\Delta \leq 0}}
\leq
n \log{\E_{X_i}\br{e^{-\pr{\sum_{j \in U} X_{ij} - \sum_{j \in V} X_{ij}}^2 / \pr{8\sigma^2}}}}.
\end{equation}
Since \(U \cap V = \varnothing\), we have:
\[
\sum_{j \in U} X_{ij} - \sum_{j \in V} X_{ij} \dequal \sum_{j \in U \cup V} X_{ij}.
\]
Therefore:
\begin{align*}
\E\br{e^{-\pr{\sum_{j \in U} X_{ij} - \sum_{j \in V} X_{ij}}^2 / \pr{8\sigma^2}}}
&=
\E\br{e^{-\pr{\sum_{j \in U \cup V} X_{ij}}^2 / \pr{8\sigma^2}}}\\
&=
\E\br{e^{-\pr{\sum_{j \in U \cup V} B_{ij} N_{ij}}^2 / \pr{8\sigma^2}}}\\
&=
\E_{B_i}\br{ \E_{N_i}\br{e^{-\pr{\sum_{j \in U \cup V} B_{ij} N_{ij}}^2 / \pr{8\sigma^2}} \, \big| \, B_{i} }}\\
&=
\E_{B_i}\br{ \E_{N_i}\br{e^{-\pr{\frac{\sum_{j \in U \cup V} B_{ij} N_{ij}}{\sqrt{\sum_{j \in U \cup V} B_{ij}}}}^2 \times \frac{\sum_{j \in U \cup V} B_{ij}}{8 \sigma^2}} \, \Bigg| \, B_{i} }}.
\end{align*}
In addition, conditionally on \(B_i\), we have:
\[
Q \coloneqq \pr{\frac{\sum_{j \in U \cup V} B_{ij} N_{ij}}{\sqrt{\sum_{j \in U \cup V} B_{ij}}}}^2
\dequal
\chi^2(1).
\]
Its MGF is:
\[
\E\br{e^{t Q} \, | \, B_i} = M_{Q |B_i}\pr{t} = \frac{1}{\sqrt{1 - 2t}}, \text{ for } t < 1/2.
\]
Hence:
\begin{align*}
\E_{B_i}\br{ \E_{N_i}\br{e^{-\pr{\frac{\sum_{j \in U \cup V} B_{ij} N_{ij}}{\sqrt{\sum_{j \in U \cup V} B_{ij}}}}^2 \times \frac{\sum_{j \in U \cup V} B_{ij}}{8 \sigma^2}} \, \Bigg| \, B_{i} }}
&=
\E_{B_i}\br{M_{Q |B_i}\pr{-\frac{\sum_{j \in U \cup V} B_{ij}}{8 \sigma^2}}}\\
&=
\E_{B_i}\br{\frac{2 \sigma}{\sqrt{4 \sigma^2 + \sum_{j \in U \cup V} B_{ij}}}}.
\end{align*}
Let \(U_{[\delta s]}\) and \(V_{[\delta s]}\) respectively denote the sets of \(\delta s\) smallest elements of \(U\) and \(V\). Note that this definition is legitimate since \(|U|=|V|=M \geq \delta s\). Since \(B_{ij} \geq 0\) for all \(j \in U \cup V\), we have:
\[
\sum_{j \in U \cup V} B_{ij}
\geq
\sum_{j \in U_{[\delta s]} \cup V_{[\delta s]}} B_{ij},
\]
and hence:
\[
\E_{B_i}\br{\frac{2 \sigma}{\sqrt{4 \sigma^2 + \sum_{j \in U \cup V} B_{ij}}}}
\leq
\E_{B_i}\br{\frac{2 \sigma}{\sqrt{4 \sigma^2 + \sum_{j \in U_{[\delta s]} \cup V_{[\delta s]}} B_{ij}}}}.
\]
Therefore, we get:
\[
\E\br{e^{-\pr{\sum_{j \in U} X_{ij} - \sum_{j \in V} X_{ij}}^2 / \pr{8\sigma^2}}}
\leq
\E_{B_i}\br{\frac{2 \sigma}{\sqrt{4 \sigma^2 + \sum_{j \in U_{[\delta s]} \cup V_{[\delta s]}} B_{ij}}}},
\]
and plugging this into \eqref{eq:logErrLeqNLogMgf} yields:
\begin{equation}\label{eq:logErrLeqNLogEB}
\log{\Pro\pr{\Delta \leq 0}}
\leq
n \log\pr{
\E_{B_i}\br{\frac{2 \sigma}{\sqrt{4 \sigma^2 + \sum_{j \in U_{[\delta s]} \cup V_{[\delta s]}} B_{ij}}}}}.
\end{equation}

Now note that, for any \(i \in [n]\):
\[
\sum_{j \in U_{[\delta s]} \cup V_{[\delta s]}} B_{ij}
\dequal
\text{Bin}\pr{2\delta s, \frac{d}{p}},
\]
In addition, since \(d = o\pr{p}\) and \(ds/p \rightarrow +\infty\), we have:

\begin{lemma}\label{lemma:AsymptoticBinomial}
For any \(i \in [n]\), the following holds: as \(p \rightarrow +\infty\),
\[
\frac{1}{\sqrt{2 \delta ds/p}}\pr{\sum_{j \in U_{[\delta s]} \cup V_{[\delta s]}} B_{ij} - 2 \delta ds/p}
\stackrel{\text{dist}}{\longrightarrow}
\mathcal{N}\pr{0,1}.
\]
\end{lemma}

\textit{Proof.} The proof is a simple adaptation of the proof of the Central Limit Theorem. See appendix \ref{section:proofOflemma:AsymptoticBinomial}.\\

Define the standardized partial sum
\[
N_p \coloneqq \frac{1}{\sqrt{2\delta ds/p}}\pr{\sum_{j \in U_{[\delta s]} \cup V_{[\delta s]}} B_{ij} - 2\delta ds/p},
\]
which is exactly the quantity appearing in Lemma~\ref{lemma:AsymptoticBinomial}, so that
\(N_p \stackrel{\text{dist}}{\longrightarrow} \calN(0,1)\) and in particular \(N_p = O(1)\). By construction,
\[
\sum_{j \in U_{[\delta s]} \cup V_{[\delta s]}} B_{ij} = 2\delta ds/p + \sqrt{2\delta ds/p}\,N_p,
\]
with no remainder term, so that
\[
V_p \coloneqq \frac{2\sigma\sqrt{ds/p}}{\sqrt{4\sigma^2 + \sum_{j \in U_{[\delta s]} \cup V_{[\delta s]}} B_{ij}}}
= \frac{2\sigma}{\sqrt{2\delta + \dfrac{4\sigma^2}{ds/p} + \sqrt{\dfrac{2\delta}{ds/p}}\,N_p}}.
\]
Since \(ds/p \rightarrow +\infty\) and \(N_p = O(1)\), the last two terms under the root vanish in probability, hence
\[
V_p \stackrel{\Pro}{\longrightarrow} \frac{2\sigma}{\sqrt{2\delta}} = \sqrt{\frac{2\sigma^2}{\delta}}.
\]
In addition, we note the following:
\begin{lemma}
\label{lemma:uniformIntegrabilityOfVp}
\(V_p\) is uniformly integrable, that is: there exists \(p' \in \N\) such that
\[
\lim_{T \rightarrow +\infty} \sup_{p \geq p'} \E\br{\abs{V_p} \indic_{\{\abs{V_p} > T\}}} = 0.
\]
\end{lemma}

\textit{Proof.} See appendix \ref{section:proofOflemma:uniformIntegrabilityOfVp}.\\

Since \(V_p \stackrel{\Pro}{\longrightarrow} \sqrt{2\sigma^2/\delta}\) and \(\pr{V_p}_p\) is uniformly integrable, the convergence holds in \(L^1\); therefore
\[
\lim_{p \rightarrow +\infty} \E\br{V_p} = \sqrt{\frac{2\sigma^2}{\delta}}.
\]
Hence, we write:
\[
\E_{B_i}\br{\frac{2 \sigma}{\sqrt{4 \sigma^2 + \sum_{j \in U_{[\delta s]} \cup V_{[\delta s]}} B_{ij}}}}
= \sqrt{\frac{2\sigma^2p}{\delta ds}} + o\pr{\pr{\frac{ds}{p}}^{-1/2}}.
\]
We conclude:
\begin{align*}
\log{\Pro\pr{\Delta \leq 0}}
&\leq
n \log\pr{
\E_{B_i}\br{\frac{2 \sigma}{\sqrt{4 \sigma^2 + \sum_{j \in U_{[\delta s]} \cup V_{[\delta s]}} B_{ij}}}}
}\\
&=
n \log\pr{
\sqrt{\frac{2\sigma^2p}{\delta ds}} + o\pr{\pr{\frac{ds}{p}}^{-1/2}}
}\\
&=
n \pr{\log\pr{\sqrt{\frac{2\sigma^2p}{\delta ds}}}
+ \log\prBig{1 + o\pr{1}}}\\
&=
n \pr{\log\pr{\sqrt{\frac{2\sigma^2p}{\delta ds}}}
+ o\pr{1}}\\
&=
n \log\pr{\sqrt{\frac{2\sigma^2p}{\delta ds}}}
+ o(n),
\end{align*}
which yields the desired result:
\[
\Pro\pr{\Delta \leq 0} \leq \pr{\frac{2\sigma^2p}{\delta ds}}^{n/2} e^{o(n)}.
\]
\end{proof}

\subsection{Proof of Corollary \ref{corollary:informationTheoreticPhaseTransition}}
\label{section:proofOfcorollary:informationTheoreticPhaseTransition}
\begin{proof}[Proof of Corollary \ref{corollary:informationTheoreticPhaseTransition}]
This proof relies on bringing together Theorem \ref{theorem:sparseRecoveryForSparseMeasurements} with the following result from Wang et al. \cite{wang2010information}.

\begin{theorem}[Necessary condition for sparse ensembles, Corollary 2 of \cite{wang2010information}]
\label{theorem:wangEtAlNecessaryCondition}
Let the measurement matrix \(X \in \R^{n \times p}\) be drawn with i.i.d. elements from the following distribution:
\begin{equation}
\label{eq:modelWangetAl}
X_{ij} =
\begin{cases}
    \calN\pr{0, \frac{1}{\gamma}}, & \text{w.p. } \gamma\\
    0 , & \text{w.p. } 1- \gamma
\end{cases},
\quad \text{ for all } i \in [n], \; j \in [p];
\end{equation}
where \(\gamma \in (0,1]\). Let \(\lambda > 0\) and
\[
\calC_{p,s}\pr{\lambda}
\coloneqq
\ac{
\beta \in \R^{p} \; \big| \;
\abs{\support{\beta}} = s, \,
\min_{i \in \support{\beta}} \abs{\beta_i} = \lambda
}.
\]
Assume that \(\sigma^2 = 1\). Then, in the regime where \(\gamma s \rightarrow +\infty\), a necessary condition for asymptotically reliable recovery over the signal class \(\calC_{p,s}\pr{\lambda}\) is given by:
\[
n > \frac{\log\binom{p}{s} - 1}{\frac{1}{2} \log\pr{1 + s \lambda^2}}.
\]
\end{theorem}
While Theorem \ref{theorem:wangEtAlNecessaryCondition} is not stated on the exact same signal space $\calC_{p,s}\pr{\lambda}$ in \cite{wang2010information} but rather on the larger:
$$
\ac{
\beta \in \R^{p} \; \big| \;
\abs{\support{\beta}} = s, \,
\min_{i \in \support{\beta}} \abs{\beta_i} \geq \lambda
},
$$
it follows directly from their result on ``restricted ensembles'' where the signal components under consideration are set exactly to $\lambda$ (see section III.A. in \cite{wang2010information}).

\subsubsection{Necessary condition}

We show that (\textit{i}) holds using Theorem \ref{theorem:wangEtAlNecessaryCondition}, but this requires adapting our problem to the framework used by Wang et al. in \cite{wang2010information}. In fact, note that the model used in Theorem \ref{theorem:wangEtAlNecessaryCondition} is different from the one we use in this paper, that we defined in \eqref{eq:model}. In their model, Wang et al. \cite{wang2010information} rescale the non-zero components of \(X\) by multiplying them by \(1/\sqrt{\gamma}\), and require that the noise variance is \(\sigma^2 = 1\). Therefore, we cannot directly use Theorem \ref{theorem:wangEtAlNecessaryCondition} in our setting. However, this difference can be fixed by a simple rescaling of our model. Note that our model defined by \eqref{eq:model}, where \(X\) follows the sparse Gaussian distribution defined in Definition \ref{def:sparseGaussianMatrix} and \(Z \sim \calN\pr{0, \sigma^2 I_n}\), can be equivalently written as:
\[
Y_0 = X_0 \beta^\star_0 + Z_0,
\]
where:
\[
Y_0 \coloneqq \frac{1}{\sigma} \, Y, \quad
X_0 \coloneqq \frac{1}{\sqrt{d/p}} \, X, \quad
\beta^\star_0 \coloneqq \frac{\sqrt{d/p}}{\sigma} \, \beta^\star,
\quad
\text{and}
\quad
Z_0 \coloneqq \frac{1}{\sigma} Z \sim \calN\pr{0, I_n}.
\]
Hence, it is a particular case of the model defined in \eqref{eq:modelWangetAl}, with:
\[
\gamma \coloneqq d/p
\quad
\text{and}
\quad
\lambda \coloneqq \frac{\sqrt{\gamma}}{\sigma}.
\]
In addition, the regime we consider of \(d = \omega\pr{p/s}\) corresponds exactly to the regime considered in Theorem \ref{theorem:wangEtAlNecessaryCondition} where \(\gamma s \rightarrow +\infty\). Therefore, using Theorem \ref{theorem:wangEtAlNecessaryCondition}, a necessary condition for an asymptotically reliable recovery of \(\beta^\star\) in the considered regime is given by:
\begin{equation}\label{eq:necesaryConditionInProofOfCorollaryPhaseTransition}
n > \frac{\log\binom{p}{s} - 1}{\frac{1}{2} \log\pr{1 + s \lambda^2}} = \frac{2\log\binom{p}{s}-2}{\log\pr{1+ds/\pr{p \sigma^2}}} = \frac{2\log{\binom{p}{s}}}{\log\pr{ds/p}} \pr{1+o\pr{1}}.
\end{equation}

\paragraph{Sublinear regime: \(s=o\pr{p}\)} Using the Corollary of Stirling:
\[
\log{\binom{p}{s}}
= s \log\pr{p/s} \pr{1 + o\pr{1}},
\]
the necessary condition \eqref{eq:necesaryConditionInProofOfCorollaryPhaseTransition} writes:
\[
n > \frac{2s\log\pr{p/s}}{\log\pr{ds/p}} \pr{1+o\pr{1}}.
\]
Let:
\[
\nInfSp \coloneqq \frac{2s\log\pr{p/s}}{\log\pr{ds/p}}.
\]
Assume there exists \(\varepsilon > 0\) such that \(n \leq \pr{1-\varepsilon}\nInfSp\). We know that, for large enough \(n,p,s,d\) we have:
\[
n \leq \pr{1-\varepsilon} \nInfSp < \frac{2s\log\pr{p/s}}{\log\pr{ds/p}} \pr{1+o\pr{1}},
\]
which contradicts the necessary condition. Therefore, it is information-theoretically impossible to ensure a reliable recovery of the support of \(\beta^\star\).\\

\paragraph{Linear regime: \(s=\alpha p, \alpha \in \pr{0,1}\).} Using the Corollary of Stirling:
\[
\log{\binom{p}{s}}
= p h\pr{\alpha} \pr{1 + o\pr{1}},
\]
the necessary condition \eqref{eq:necesaryConditionInProofOfCorollaryPhaseTransition} writes:
\[
n > \frac{2 h\pr{\alpha} p}{\log{d}} \pr{1+o\pr{1}}.
\]
Let:
\[
\nInfSp \coloneqq \frac{
2 h\pr{\alpha} p}{\log{d}}.
\]
Similarly to above, we conclude that if there exists \(\varepsilon > 0\) such that \(n \leq \pr{1-\varepsilon} \nInfSp\) then it is information-theoretically impossible to ensure a reliable recovery of the support of \(\beta^\star\).

\subsubsection{Sufficient condition}

We show that (\textit{ii}) holds using Theorem \ref{theorem:sparseRecoveryForSparseMeasurements}.

\paragraph{Sublinear regime: \(s=o\pr{p}\).}
Let \(\delta > 0\) and:
\[
n^\star_{\text{SP}} \coloneqq \frac{2 s \log\pr{p/s}}{\log\pr{ds/p} + \log\pr{\delta/(2\sigma^2)}}. 
\]
Note that:
\[
\nInfSp = \frac{2s\log\pr{p/s}}{\log\pr{ds/p}} = n^\star_{\text{SP}} \pr{1+o\pr{1}}.
\]
Assume there exists \(\varepsilon\) such that \(n \geq \pr{1+\varepsilon} \nInfSp\). Then:
\[
n \geq \pr{1+\varepsilon} \pr{1+o\pr{1}} n^\star_{\text{SP}} = \pr{1+\varepsilon+o\pr{1}} n^\star_{\text{SP}} \geq \big(1+\varepsilon/2\big) n^\star_{\text{SP}},
\]
for \(n,p,s,d\) large enough. Using Theorem \ref{theorem:sparseRecoveryForSparseMeasurements}, we have:
\[
\Pro_{X,Z}\pr{\frac{1}{2s} \abs{\symdiff{\support{\hat{\beta}}}{\support{\beta^\star}}} < \delta}
\geq
1 - \exp\Big(- \varepsilon s \log\pr{p/s} + o\,\big(s \log\pr{p/s}\big)\Big).
\]
Therefore, we obtain:
\[
\Pro_{X,Z}\pr{\frac{1}{2s} \abs{\symdiff{\support{\hat{\beta}}}{\support{\beta^\star}}} < \delta}
\longrightarrow 1,
\]
as \(n,p,s,d \rightarrow +\infty\). Since this holds for all \(\delta > 0\), we conclude:
\[
\frac{\abs{\symdiff{\support{\hat{\beta}}}{\support{\beta^\star}}}}{2s} \longrightarrow 0,
\]
in probability, as \(n,p,s,d \rightarrow +\infty\).

\paragraph{Linear regime: \(s = \alpha p, \alpha \in \pr{0,1}\).}
Let \(\delta > 0\) and:
\[
n^\star_{\text{SP}} \coloneqq \frac{2 h\pr{\alpha} p}{\log{d} + \log\pr{\delta \alpha/(2\sigma^2)}}. 
\]
Note that:
\[
\nInfSp = \frac{2 h\pr{\alpha} p}{\log{d}} = n^\star_{\text{SP}} \pr{1+o\pr{1}}.
\]
Assume there exists \(\varepsilon\) such that \(n \geq \pr{1+\varepsilon} \nInfSp\). Then:
\[
n \geq \pr{1+\varepsilon} \pr{1+o\pr{1}} n^\star_{\text{SP}} = \pr{1+\varepsilon+o\pr{1}} n^\star_{\text{SP}} \geq \big(1+\varepsilon/2\big) n^\star_{\text{SP}},
\]
for \(n,p,s,d\) large enough. Using Theorem \ref{theorem:sparseRecoveryForSparseMeasurements}, we have:
\[
\Pro_{X,Z}\pr{\frac{1}{2s} \abs{\symdiff{\support{\hat{\beta}}}{\support{\beta^\star}}} < \delta}
\geq
1 - \exp\Big(- \varepsilon h\pr{\alpha} p + o\pr{p}\Big).
\]
Therefore, we obtain:
\[
\Pro_{X,Z}\pr{\frac{1}{2s} \abs{\symdiff{\support{\hat{\beta}}}{\support{\beta^\star}}} < \delta}
\longrightarrow 1,
\]
as \(n,p,s,d \rightarrow +\infty\). Since this holds for all \(\delta > 0\), we conclude:
\[
\frac{\abs{\symdiff{\support{\hat{\beta}}}{\support{\beta^\star}}}}{2s} \longrightarrow 0,
\]
in probability, as \(n,p,s,d \rightarrow +\infty\).
\end{proof}

\subsection{Proof of Example \ref{example:measurementSparsitySamplingComplexityTradeoff}}
\label{section:proofOfExample:measurementSparsitySamplingComplexityTradeoff}
\begin{proof}[Proof of Example \ref{example:measurementSparsitySamplingComplexityTradeoff}]
We have \(d = \min\pr{p^{o\pr{1}}, \varphi^{-1}\pr{o\pr{\varphi\pr{p}}}}\), hence:
\[
\begin{cases}
\log{d}/\log{p} = o\pr{1}\\
\varphi\pr{d}/\varphi\pr{p} = o\pr{1}
\end{cases}.
\]
In addition:
\[
n_1 = \nInf = \Theta\prbig{p/\log{p}},
\quad
n_2 = \nInfSp = \Theta\prbig{p/\log{d}}.
\]
Therefore, we have:
\[
\frac{n_2}{n_1}
= \frac{\Theta\pr{p/\log{d}}}{\Theta\pr{p/\log{p}}}
= \Theta\pr{\frac{\log{p}}{\log{d}}} = \omega\pr{1},
\]
On one hand, the number of samples required for reliable recovery is better in the dense case:
\begin{equation}\label{eq:complexityTradeoffExampleN_2>N_1}
n_1
= \Theta\prbig{p/\log{p}}
= o\prbig{p/\log{d}}
= o\pr{n_2}.
\end{equation}
On the other hand, the computational cost of recovering the support is better in the sparse case. In fact, matrix-vector multiplications are made easier by sparsity: in the dense case, multiplying \(X_1\) with a vector in \(\R^p\) costs:
\[
n_1 p = \Theta\prbig{p^2/\log{p}}
\]
real number multiplications, while multiplying \(X_2\) with a vector in \(\R^p\) costs
\begin{equation}\label{eq:complexityTradeoffExampleN_2d<N_1p}
n_2 d
= \Theta\prbig{pd/\log{d}}
= p \Theta\prbig{\varphi\pr{d}}
= p o\prbig{\varphi\pr{p}}
= o\prbig{p^2/\log{p}}
= o\pr{n_1 p},
\end{equation}
real number multiplications. This highlights the trade-off between sampling complexity and computational cost.
\end{proof}

\section{Sparse Recovery via Active Sparsification: Proof of Theorem \ref{theorem:sparsificationLinear}}
\label{section:proofOftheorem:sparsificationLinear}
We assume, to avoid notational distractions, that \(\alpha p\) and \(\psi p\) are integers; since standard rounding only affects lower-order terms.

Let \(S^\star\coloneqq \support{\beta^\star}\) and let
\[
    \calS \coloneqq \ac{S\subset[p]: |S|=s}.
\]
For \(S\in\calS\), define
\[
    L(S)\coloneqq \norm{\Tilde{Y}-\Tilde{X}\indic_S}_2^2.
\]
For a competing support \(S\), write
\[
    A=A(S)\coloneqq S^\star\setminus S,
    \qquad
    B=B(S)\coloneqq S\setminus S^\star,
    \qquad
    C=C(S)\coloneqq S^\star\cap S,
\]
and let
\[
    M=M(S)\coloneqq |A|=|B|=\frac12\abs{S\triangle S^\star}.
\]
We put \(\eta\coloneqq M/s\in[0,1]\). For a fixed support \(S\), set
\[
    \Delta_S \coloneqq L(S)-L(S^\star).
\]

Before stating and proving the main large-deviation estimate, we outline the strategy of the proof.

The argument follows the same general outline as Theorem~\ref{theorem:sparseRecoveryForSparseMeasurements}: a Chernoff bound on the row-wise loss difference \(\Delta_S\), followed by a union bound over supports \(S\) with \(\lvert S\triangle S^\star\rvert\geq 2\delta s\). The technical work, however, is substantially more involved, because \(\widetilde{Y}\) is not a noisy projection of the true signal through \(\widetilde{X}\) but a rescaling of the original observations \(Y\). As a result, \(\Delta_S\) does not decompose as cleanly as in the intrinsically-sparse setting, and its row moment generating function depends on the realization of the sparsification mask \(B_1\).

The strategy is to compute the row MGF \emph{conditionally on \(B_1\)}. After integrating out the noise \(Z_1\), the conditional MGF takes the form \(\E[\exp(U_\theta V_\theta)\mid B_1]\) for a centered Gaussian pair \((U_\theta, V_\theta)\) whose variances and covariance are explicit polynomials in the mask sums \(x_A(B_1), x_B(B_1), x_C(B_1)\) over the index sets \(A, B, C\) defined above (Lemma~\ref{lemma:GaussianProductMGFNew}). This expression is finite only when a discriminant \(D_p(\theta, b) = (1 - c_p(\theta,b))^2 - a_p(\theta,b)b_p(\theta,b)\) is positive.

A natural choice of Chernoff parameter \(\theta^\star\) makes the limiting MGF collapse to a clean closed form via an algebraic cancellation (see the calculation in the proof of Lemma~\ref{lemma:RowMGFLimitSparsification}, where the \(\eta\psi^2\) and \(2 - \eta - 2\psi(1-\eta)\) terms combine into \(-T_\eta\)). The drawback is that \(D_p(\theta^\star, b)\) vanishes on a set of masks \(b\) of exponentially small probability, so directly evaluating the bound at \(\theta^\star\) would require a uniform-integrability hypothesis to justify passing to the limit. We sidestep this by evaluating the Chernoff bound at the shrunken parameter \(\theta_{p,\lambda} \coloneqq \lambda\theta^\star\) for some fixed \(\lambda\in(0,1)\). The factor \(\lambda < 1\) keeps \(D_p(\theta_{p,\lambda}, b)\) bounded below by a positive constant uniformly over all masks \(b\) (Lemma~\ref{lemma:UniformFinitenessSparsification}), hence yielding the limit by bounded convergence.

The cost of this shrinkage is a degraded MGF limit: instead of \((1+C)^{-1/2}\), we obtain \((1+(2\lambda-\lambda^2)C)^{-1/2}\) (where \(C>0\) is a constant depending on the problem parameters, whose explicit expression we omit here for readability). Since \(2\lambda - \lambda^2 \to 1\) as \(\lambda\to 1^-\), the slack \(\varepsilon\) in the sample-size assumption absorbs this loss.\\

We now state the main large-deviation estimate, uniform over all supports whose relative error is at least \(\delta\).

\begin{proposition}\label{proposition:sparsificationLargeDeviationBound}
Fix \(\alpha,\delta\in(0,1)\) and \(\lambda\in(0,1)\). There exists \(\psi_0=\psi_0(\alpha,\delta,\lambda)>0\) such that, for every fixed \(\psi\in(0,\psi_0)\), uniformly over all \(S\in\calS\) with \(\eta \coloneqq M(S)/s \geq \delta\),
\begin{equation}\label{eq:sparsificationLDlambda}
    \Pro\pr{\Delta_S\leq 0}
    \leq
    \pr{1+(2\lambda-\lambda^2)\eta\psi C^\star(\eta)}^{-n/2} e^{o(n)},
\end{equation}
where
\begin{equation}\label{eq:CstarSparsificationProof}
    C^\star(\eta)
    \coloneqq
    \frac{\psi}{\pr{1-\psi}\pr{2-\eta\pr{1-\psi}}}.
\end{equation}
The \(o(n)\) term is uniform in \(S\) and in \(\eta\in[\delta,1]\).
\end{proposition}

Before proving Proposition \ref{proposition:sparsificationLargeDeviationBound}, we show how it implies Theorem \ref{theorem:sparsificationLinear}.

\begin{proof}[Proof of Theorem \ref{theorem:sparsificationLinear}]
Fix \(\varepsilon>0\). We construct \(\lambda\) and \(\psi_0\) below; \(\lambda\) will depend only on \(\varepsilon\) and \(\psi_0\) only on \((\alpha,\delta,\lambda)\), so the resulting \(\psi_0\) depends only on \((\alpha,\delta,\varepsilon)\), as the statement requires.

\emph{Step 1: choose \(\lambda\).} Pick \(\lambda\in(0,1)\) sufficiently close to one so that
\begin{equation}\label{eq:lambdaChoiceMu}
    (1+\varepsilon)(2\lambda-\lambda^2)>1.
\end{equation}
Such \(\lambda\) clearly exists.

\emph{Step 2: choose \(\psi_0\).} The function \(g(x)\coloneqq (1+\varepsilon)\log(1+(2\lambda-\lambda^2)x)-\log(1+x)\) satisfies \(g(0)=0\) and for any \(x \geq 0\) we have by \eqref{eq:lambdaChoiceMu}:
\[
g'\pr{x} = \frac{\pr{1+\varepsilon} \pr{2\lambda-\lambda^2}}{1+\pr{2\lambda-\lambda^2} x} - \frac{1}{1+x} > 0,
\]
therefore $g\pr{x} > 0$ for all $x \geq 0$. Hence, for every $\psi \in \pr{0,1}$ we have:
\begin{equation}\label{eq:lambdaChoiceSparsification}
    (1+\varepsilon)\log\pr{1+(2\lambda-\lambda^2)\delta\psi C^\star(\delta)}
    >
    \log\pr{1+\delta\psi C^\star(\delta)}.
\end{equation}
Let \(\psi_0\) be the threshold supplied by Proposition \ref{proposition:sparsificationLargeDeviationBound} for the chosen \(\lambda\). Fix any \(\psi\in(0,\psi_0)\) and assume \(n\geq(1+\varepsilon)n^\star_{\text{SP}}\).

\emph{Step 3: union bound.} By Proposition \ref{proposition:sparsificationLargeDeviationBound}, for every support \(S\) with \(M(S)\geq\delta s\),
\[
    \Pro\pr{\Delta_S\leq 0}
    \leq
    \pr{1+(2\lambda-\lambda^2)\eta\psi C^\star(\eta)}^{-n/2} e^{o(n)}.
\]
The function
\[
    \eta\longmapsto \eta C^\star(\eta)
    =
    \frac{\eta\psi}{\pr{1-\psi}\pr{2-\eta\pr{1-\psi}}}
\]
is increasing on \([0,1]\). Hence, for \( \eta \geq \delta \):
\[
    \Pro\pr{\Delta_S\leq 0}
    \leq
    \pr{1+(2\lambda-\lambda^2)\delta\psi C^\star(\delta)}^{-n/2}e^{o(n)}.
\]
Using the union bound over all supports of cardinality \(s\),
\begin{align*}
&\Pro\pr{\abs{\support{\hat\beta}\triangle S^\star}\geq 2\delta s} \\
&\qquad\leq
\sum_{S\in\calS\,:\,M(S)\geq\delta s}\Pro\pr{\Delta_S\leq 0} \\
&\qquad\leq
\binom{p}{s}
\pr{1+(2\lambda-\lambda^2)\delta\psi C^\star(\delta)}^{-n/2}e^{o(n)}.
\end{align*}
Since \(s=\alpha p\), Stirling's formula gives \(\log\binom{p}{s}=h(\alpha)p+o(p)\). Moreover, by the definition of \(n^\star_{\text{SP}}\),
\[
    h(\alpha)p
    =
    \frac{n^\star_{\text{SP}}}{2}
    \log\pr{1+\delta\psi C^\star(\delta)}.
\]
Note that \(o(p)=o(n)\) since \(n=\Theta(p)\) for fixed \(\psi\). Therefore
\begin{align*}
\log \Pro\pr{\abs{\support{\hat\beta}\triangle S^\star}\geq 2\delta s}
&\leq
h(\alpha)p
-\frac{n}{2}
\log\pr{1+(2\lambda-\lambda^2)\delta\psi C^\star(\delta)}
+o(n)\\
&\leq
\frac{n^\star_{\text{SP}}}{2}
\Big[
\log\pr{1+\delta\psi C^\star(\delta)} \\
&\qquad\qquad
-(1+\varepsilon)
\log\pr{1+(2\lambda-\lambda^2)\delta\psi C^\star(\delta)}
\Big]
+o(n).
\end{align*}
By \eqref{eq:lambdaChoiceSparsification}, the bracketed term is a strictly negative constant. Since \(n^\star_{\text{SP}}=\Theta(p)\) for fixed \(\psi\), the right-hand side tends to \(-\infty\). This proves
\[
    \Pro\pr{\abs{\support{\hat\beta}\triangle S^\star}<2\delta s}\longrightarrow 1.
\]
\end{proof}

It remains to prove Proposition \ref{proposition:sparsificationLargeDeviationBound}.

\subsection{Conditional Chernoff transform}

Fix \(S\in\calS\) and write \(A,B,C,M,\eta\) as above. We have:
\begin{align*}
\Delta_S
&= L(S) - L(S^\star)\\
&= \norm{\frac{d}{p} X \indic_{S^\star} - \Tilde{X} \indic_{S} + \frac{d}{p} Z}_2^2 - \norm{\pr{\frac{d}{p} X - \Tilde{X}} \indic_{S^\star} + \frac{d}{p} Z}_2^2\\
&= \norm{\Tilde{X} \indic_{S}}_2^2 - \norm{\Tilde{X} \indic_{S^\star}}_2^2 + 2 \; \frac{d}{p} \; \inner{X \indic_{S^\star} + Z}{\Tilde{X} \pr{\indic_{S^\star} - \indic_{S}}}\\
&= \sum_{i=1}^{n} \inner{\Tilde{X}_i}{\indic_{S}}^2 - \sum_{i=1}^{n} \inner{\Tilde{X}_i}{\indic_{S^\star}}^2 + 2 \; \frac{d}{p} \; \sum_{i = 1}^{n} \pr{\inner{X_i}{\indic_{S^\star}} + Z_i} \pr{\inner{\Tilde{X}_i}{\indic_{S^\star}} - \inner{\Tilde{X}_i}{\indic_{S}}}\\
&= \sum_{i=1}^{n} \pr{ \inner{\Tilde{X}_i}{\indic_{S}}^2 - \inner{\Tilde{X}_i}{\indic_{S^\star}}^2 + 2 \; \psi \; \pr{\inner{X_i}{\indic_{S^\star}} + Z_i} \pr{\inner{\Tilde{X}_i}{\indic_{S^\star}} - \inner{\Tilde{X}_i}{\indic_{S}}} }.
\end{align*}
For each row \(i\), let
\[
    \Delta_i
    \coloneqq
    \inner{\Tilde X_i}{\indic_S}^2
    -
    \inner{\Tilde X_i}{\indic_{S^\star}}^2
    +2\psi\pr{\inner{X_i}{\indic_{S^\star}}+Z_i}
    \pr{\inner{\Tilde X_i}{\indic_{S^\star}}-\inner{\Tilde X_i}{\indic_S}}.
\]
Then \(\Delta_S=\sum_{i=1}^n\Delta_i\), and the \(\Delta_i\)'s are i.i.d. For \(\theta>0\), Chernoff's bound gives
\begin{equation}\label{eq:ChernoffBoundSparsificationNew}
    \Pro\pr{\Delta_S\leq 0}
    \leq
    \pr{\E\br{e^{-\theta\Delta_1}}}^n,
\end{equation}
for all $\theta > 0$. We compute the row moment generating function. Let \(B_1=(B_{1j})_{j\in[p]}\) be the first row of the sparsification mask. Conditionally on \(B_1\), integrating first over \(Z_1\sim\calN(0,\sigma^2)\) yields
\begin{equation}\label{eq:conditionalMGFPreUV}
\E\br{e^{-\theta\Delta_1}\mid B_1}
=\E\br{e^{U_\theta V_\theta}\mid B_1},
\end{equation}
where we define the centered jointly Gaussian variables
\begin{align*}
U_\theta
&\coloneqq
\theta\pr{
\sum_{j\in A} B_{1j}X_{1j}
-
\sum_{j\in B} B_{1j}X_{1j}},\\
V_\theta
&\coloneqq
\sum_{j\in A}\pr{\pr{1+\gamma(\theta)}B_{1j}-2\psi}X_{1j}
+
\sum_{j\in B}\pr{1-\gamma(\theta)}B_{1j}X_{1j}\\
&\qquad
+2\sum_{j\in C}\pr{B_{1j}-\psi}X_{1j},
\end{align*}
with
\begin{equation}\label{equation:definitionGamma}
    \gamma(\theta)\coloneqq 2\psi^2\sigma^2\theta.
\end{equation}
The identity \eqref{eq:conditionalMGFPreUV}, with the stated forms of \(U_\theta\) and \(V_\theta\), follows from a direct expansion of \(\Delta_1\) and a Gaussian integration over \(Z_1\); the computation is carried out in Appendix \ref{section:conditionalChernoffExpansion}.

We now introduce the following lemma to characterize the entity in \eqref{eq:conditionalMGFPreUV}.

\begin{lemma}\label{lemma:GaussianProductMGFNew}
Let \((U,V)\) be a centered bivariate Gaussian vector. Put
\[
    a \coloneqq \var(U),\qquad b \coloneqq \var(V),\qquad c \coloneqq \cov(U,V).
\]
If
\[
    D\coloneqq (1-c)^2-ab>0,
\]
then
\[
    \E\br{e^{UV}}=D^{-1/2}.
\]
If \(D\leq0\), then the expectation is infinite.
\end{lemma}

The proof Lemma \ref{lemma:GaussianProductMGFNew} follows from a simple change of variable: see appendix \ref{section:proofOflemma:GaussianProductMGFNew}.

For a deterministic mask realization \(b=(b_j)_{j\in[p]}\in\{0,1\}^p\), define
\begin{align*}
    x_A(b)&\coloneqq \sum_{j\in A}b_j,
    &x_B(b)&\coloneqq \sum_{j\in B}b_j,
    &x_C(b)&\coloneqq \sum_{j\in C}b_j.
\end{align*}
Conditionally on \(B_1=b\), we know that since the \(X_{1j}\) are independent \(\calN(0,1)\), \((U_\theta,V_\theta)\) is centered and jointly Gaussian. A direct computation (Appendix \ref{section:conditionalMomentsUV}) yields its variances and covariance:
\begin{align}
    a_p(\theta,b)
    &\coloneqq \var(U_\theta\mid B_1=b)
    =\theta^2\pr{x_A(b)+x_B(b)},\label{eq:apSparsification}\\
    b_p(\theta,b)
    &\coloneqq \var(V_\theta\mid B_1=b)\notag\\
    &=4\psi^2s
    +\pr{\pr{1+\gamma(\theta)}^2-4\psi\pr{1+\gamma(\theta)}}x_A(b)
    +\pr{1-\gamma(\theta)}^2x_B(b)\notag\\
    &\qquad +4(1-2\psi)x_C(b),\label{eq:bpSparsification}\\
    c_p(\theta,b)
    &\coloneqq \cov(U_\theta,V_\theta\mid B_1=b)\notag\\
    &=\theta\br{\pr{1+\gamma(\theta)-2\psi}x_A(b)-\pr{1-\gamma(\theta)}x_B(b)}.
    \label{eq:cpSparsification}
\end{align}
Put
\begin{equation}\label{eq:DpSparsification}
    D_p(\theta,b)
    \coloneqq
    \pr{1-c_p(\theta,b)}^2-a_p(\theta,b)b_p(\theta,b).
\end{equation}
Whenever \(D_p(\theta,b)>0\), Lemma \ref{lemma:GaussianProductMGFNew} gives
\begin{equation}\label{eq:rowMGFWithDp}
    \E\br{e^{-\theta\Delta_1}\mid B_1=b}=D_p(\theta,b)^{-1/2}.
\end{equation}

\subsection{A regularized Chernoff parameter}

For \(\lambda\in(0,1)\), define
\begin{equation}\label{eq:thetaLambdaEta}
    \theta_{p,\lambda}(\eta)
    \coloneqq
    \frac{\lambda C^\star(\eta)}{2\alpha\psi p}
    =
    \frac{\lambda}{2\alpha\pr{1-\psi}\pr{2-\eta\pr{1-\psi}}}\cdot\frac1p.
\end{equation}
The shrinkage factor \(\lambda<1\) keeps the Gaussian product MGF uniformly inside its domain of finiteness.

\begin{lemma}\label{lemma:UniformFinitenessSparsification}
Fix \(\alpha,\delta\in(0,1)\) and \(\lambda\in(0,1)\). There exist \(\psi_0>0\), \(p_0\in\N\), and \(\kappa>0\) such that, for every \(\psi\in(0,\psi_0)\), every \(p\geq p_0\), every \(\eta\in[\delta,1]\), and every mask \(b\in\{0,1\}^p\),
\[
    D_p\pr{\theta_{p,\lambda}(\eta),b}\geq \kappa.
\]
Consequently,
\[
    \E\br{e^{-\theta_{p,\lambda}(\eta)\Delta_1}\mid B_1=b}
    \leq \kappa^{-1/2}
\]
for all such \(b\).
\end{lemma}

\begin{proof}
Write
\[
    u\coloneqq \frac{x_A(b)}p,
    \qquad
    v\coloneqq \frac{x_B(b)}p,
    \qquad
    w\coloneqq \frac{x_C(b)}p.
\]
Recall that \(\abs{A} = \abs{B} = \eta s\) and \(\abs{C} = \pr{1-\eta} s\). Then \(0\leq u,v\leq\eta\alpha\) and \(0\leq w\leq (1-
\eta)\alpha\). Let
\[
    k_{\lambda,\eta,\psi}
    \coloneqq
    \frac{\lambda}{2\alpha\pr{1-\psi}\pr{2-\eta\pr{1-\psi}}},
    \quad
    \text{so that}
    \quad
    \theta_{p,\lambda}(\eta)=k_{\lambda,\eta,\psi}/p.
\]
By \eqref{equation:definitionGamma} and \eqref{eq:thetaLambdaEta}, we have \(\gamma(\theta_{p,\lambda}(\eta))=O(p^{-1})\). Hence, uniformly over \(\eta\in[\delta,1]\), \(\psi\) in \([0,1)\), and all masks \(b\), the expression \(D_p\pr{\theta_{p,\lambda}(\eta),b}\) differs by \(o(1)\) from
\begin{align*}
D_{\psi}(u,v,w;\eta)
&\coloneqq
\pr{1-k_{\lambda,\eta,\psi}\br{(1-2\psi)u-v}}^2\\
&
\quad
-k_{\lambda,\eta,\psi}^2(u+v)
\br{4\psi^2\alpha+(1-4\psi)u+v+4(1-2\psi)w}.
\end{align*}

We first lower-bound the limiting expression at \(\psi=0\). We have
\[
    k_{\lambda,
\eta,0}=\frac{\lambda}{2\alpha(2-\eta)}.
\]
For \(\psi=0\),
\begin{align*}
D_0(u,v,w;\eta)
&=
\pr{1-k_{\lambda,\eta,0}(u-v)}^2
-k_{\lambda,\eta,0}^2(u+v)(u+v+4w)\\
&=
1-2k_{\lambda,\eta,0}(u-v)
-4k_{\lambda,\eta,0}^2\br{uv+w(u+v)}.
\end{align*}
The right-hand side is decreasing in \(w\), so its minimum over the allowed interval for \(w\) is attained at \(w=(1-\eta)\alpha\). With this value of \(w\), the derivative in \(v\) equals
\[
    2k_{\lambda,\eta,0}
    -4k_{\lambda,\eta,0}^2\br{u+(1-
\eta)\alpha}
    =2k_{\lambda,\eta,0}
    \br{1-2k_{\lambda,\eta,0}\br{u+(1-
\eta)\alpha}}
    \geq 0,
\]
because \(u+(1-\eta)\alpha\leq\alpha\) (recall that \(u \leq \eta \alpha\)) and \(2k_{\lambda,\eta,0}\alpha=\lambda/(2-
\eta)<1\). Hence the minimum is attained at \(v=0\). The resulting expression,
\[
D_0\pr{u, 0, \pr{1-\eta} \alpha ; \eta}
=
1 - 2 k_{\lambda,\eta,0} u - 4 k_{\lambda,\eta,0}^2 \pr{1-\eta} \alpha u,
\]
is decreasing in \(u\), and hence the minimum is attained at \(u=
\eta\alpha\). Therefore
\begin{align*}
D_0(u,v,w;\eta)
&\geq
1-\eta\alpha\pr{2k_{\lambda,\eta,0}+4k_{\lambda,\eta,0}^2(1-
\eta)
\alpha}\\
&=
1 - \frac{2 \lambda \eta \alpha}{2 \alpha \pr{2-\eta}}
- \frac{4 \eta \alpha \pr{1-\eta} \alpha \lambda^2}{4 \alpha^2 \pr{2-\eta}^2}
\\
&=
1-
\lambda\frac{
\eta}{2-
\eta}
-
\lambda^2\frac{
\eta(1-
\eta)}{(2-
\eta)^2}\\
&\geq 1-
\lambda,
\end{align*}
where the last inequality holds because \(\lambda\leq1\) and
\[
    \frac{
\eta}{2-
\eta}+\frac{
\eta(1-
\eta)}{(2-
\eta)^2}
    \leq 1,
    \quad \forall \,
\eta\in[0,1].
\]
By continuity of \(D_\psi\) in \(\psi \in [0,1)\), uniformly over the same compact domain, there exists \(\psi_0>0\) such that
\[
    D_\psi(u,v,w;\eta)
    \geq
    \frac{1-
\lambda}{2}
\]
for all \(\psi\in(0,\psi_0)\). The uniform \(o(1)\) approximation between \(D_p\) and \(D_\psi\) then gives the claim, after setting \(\kappa \coloneqq (1-\lambda)/4\), and \(p_0\) large enough (recall that \(D_p\pr{\theta_{p,\lambda}(\eta),b}\) and  \(D_{\psi}(u,v,w;\eta)\) differ by \(o\pr{1}\)).
\end{proof}

\begin{lemma}\label{lemma:RowMGFLimitSparsification}
Fix \(\alpha,
\delta\in(0,1)\) and \(\lambda\in(0,1)\), and take \(\psi>0\) small enough for Lemma \ref{lemma:UniformFinitenessSparsification} to hold. Then we have, uniformly over \(\eta\in[\delta,1]\),
\begin{equation}\label{eq:rowMGFLimitSparsification}
    \E\br{e^{-\theta_{p,\lambda}(\eta)
\Delta_1}}
    =
    \pr{1+(2\lambda-
\lambda^2)
\eta\psi C^\star(
\eta)}^{-1/2}+o(1).
\end{equation}
\end{lemma}

\begin{proof}
Let
\[
    U_p\coloneqq\frac{x_A(B_1)}p,
    \qquad
    V_p\coloneqq\frac{x_B(B_1)}p,
    \qquad
    W_p\coloneqq\frac{x_C(B_1)}p.
\]
Since \(|A|=|B|=\eta\alpha p\) and \(|C|=(1-\eta)\alpha p\), the three variables are sums of i.i.d.\ Bernoulli\((\psi)\) terms divided by \(p\). For every \(\rho>0\), Hoeffding's inequality applied to \(S_A:=\sum_{j\in A}B_{1j}\) gives
\begin{align*}
    \mathbb{P}\pr{|U_p-\eta\alpha\psi|>\rho}
    &=\mathbb{P}\pr{|S_A-\eta\alpha p\psi|>\rho p}\\
    &\le 2\exp\pr{-\frac{2\rho^2 p^2}{|A|}}\\
    &=2\exp\pr{-\frac{2\rho^2 p}{\eta\alpha}}\\
    &\le 2\exp\pr{-\frac{2\rho^2 p}{\alpha}},
\end{align*}
where the last inequality uses \(\eta \leq 1\). The same bound holds for \(V_p\). For \(W_p\), the analogous computation with \(|C|=(1-\eta)\alpha p\) gives, for \(\eta\in[\delta,1)\),
\[
    \mathbb{P}\pr{|W_p-(1-\eta)\alpha\psi|>\rho}
    \le 2\exp\pr{-\frac{2p\rho^2}{(1-\eta)\alpha}}
    \le 2\exp\pr{-\frac{2p\rho^2}{(1-\delta)\alpha}};
\]
for \(\eta=1\), \(W_p\equiv 0=(1-\eta)\alpha\psi\) and the probability vanishes. Each right-hand side is independent of \(\eta\) and tends to zero as \(p\to\infty\), so
\[
    \sup_{\eta\in[\delta,1]}\mathbb{P}\pr{|U_p-\eta\alpha\psi|>\rho}\to 0,
\]
and analogously for \(V_p,W_p\). Hence
\[
    \pr{U_p,V_p,W_p}\longrightarrow\pr{\eta\alpha\psi,\eta\alpha\psi,(1-\eta)\alpha\psi}
\]
in probability, uniformly in \(\eta\in[\delta,1]\).\\

Let
\[
    F_p(u,v,w;\eta)
    \coloneqq
    D_p\pr{\theta_{p,\lambda}(\eta),b}^{-1/2},
\]
where \(u=x_A(b)/p\), \(v=x_B(b)/p\), and \(w=x_C(b)/p\). By Lemma \ref{lemma:UniformFinitenessSparsification}, \(0 \leq F_p \leq \kappa^{-1/2}\) for all masks and all \(\eta \in [\delta, 1]\). Define the compact set
\[
    K\coloneqq
    \ac{
    (u,v,w,\eta) :
    0 \leq u,v \leq \eta\alpha,
    \;
    0 \leq w \leq \pr{1-\eta}\alpha,
    \;
    \eta \in [\delta,1]
    }.
\]
On \(K\), the polynomial formula \eqref{eq:DpSparsification} together with \(\theta_{p,\lambda}(\eta)=O(p^{-1})\) and \(\gamma(\theta_{p,\lambda})=O(p^{-1})\) (uniformly in \(\eta\)) gives \(D_p\to D_\infty\) uniformly, where \(D_\infty(u,v,w;\eta)\) is the polynomial obtained by setting \(\gamma=0\) in \eqref{eq:DpSparsification}. Combined with the uniform lower bound \(D_p\geq\kappa>0\) and the Lipschitz continuity of \(x\mapsto x^{-1/2}\) on \([\kappa,\infty)\), this yields
\[
    \sup_{(u,v,w,\eta)\in K}
    \abs{F_p(u,v,w;\eta)-F_\infty(u,v,w;\eta)}
    \longrightarrow 0,
\]
where \(F_\infty\coloneqq D_\infty^{-1/2}\). Combining this uniform convergence with the concentration \(\pr{U_p,V_p,W_p}\to\pr{\eta\alpha\psi,\eta\alpha\psi,(1-\eta)\alpha\psi}\) in probability (uniformly in \(\eta\)) and the uniform continuity of \(F_\infty\) on \(K\), we obtain
\[
    F_p\pr{U_p,V_p,W_p;\eta}
    \longrightarrow
    F_\infty\pr{\eta\alpha\psi,\eta\alpha\psi,(1-\eta)\alpha\psi;\eta}
\]
in probability, uniformly in \(\eta\in[\delta,1]\). Since \(F_p\leq\kappa^{-1/2}\), bounded convergence yields
\[
    \sup_{\eta\in[\delta,1]}
    \abs{
    \E_B\br{D_p\pr{\theta_{p,\lambda}(\eta),B_1}^{-1/2}}
    -
    F_\infty\pr{\eta\alpha\psi,\eta\alpha\psi,(1-\eta)\alpha\psi;\eta}}
    \longrightarrow 0.
\]

Substituting
\[
    u=v=
\eta\alpha\psi,
    \quad
    w=(1-
\eta)
\alpha\psi,
    \quad
    \theta_{p,
\lambda}(
\eta)=
\frac{\lambda C^\star(
\eta)}{2\alpha\psi p},
\]
into \eqref{eq:apSparsification}, \eqref{eq:bpSparsification} and \eqref{eq:cpSparsification}; and using \(\gamma(\theta_{p,
\lambda})=o(1)\), we obtain
\begin{align*}
    c_p
    &\longrightarrow
    -\lambda 
\eta\psi C^\star(
\eta),\\
    a_p b_p
    &\longrightarrow
    \lambda^2 
\eta\psi C^\star(
\eta)
    \pr{1+
\eta\psi C^\star(
\eta)},
\end{align*}
which yields:
\begin{align*}
    \pr{1-c_p}^2-a_p b_p
    &\longrightarrow
    1+(2\lambda-
\lambda^2)
\eta\psi C^\star(
\eta).
\end{align*}
We spell out the cancellation in detail. Let
\[
    T_\eta\coloneqq
    \pr{1-
\psi}\pr{2-
\eta\pr{1-
\psi}},
    \quad \text{so that} \quad
    C^\star(
\eta)=\psi/T_\eta.
\]
At the mean mask profile,
\begin{align*}
c_p
& \longrightarrow
-2k_{\lambda,
\eta,
\psi}
\eta\alpha\psi^2
    =-
\lambda
\eta\psi C^\star(
\eta),\\
a_p b_p
& \longrightarrow
4k_{\lambda,
\eta,
\psi}^2
    \eta\alpha^2\psi^2
    \br{2-
\eta-2\psi(1-
\eta)}.
\end{align*}
Thus
\begin{align*}
\lim_{p\to\infty}D_p
&=
\pr{1+
\lambda
\eta\psi C^\star(
\eta)}^2
-
4k_{\lambda,
\eta,
\psi}^2
\eta\alpha^2\psi^2
\br{2-
\eta-2\psi(1-
\eta)}\\
&=
1+2\lambda
\eta\psi C^\star(
\eta)
+
\lambda^2
\eta^2\psi^2C^\star(
\eta)^2\\
&\qquad
-
\lambda^2
\eta C^\star(
\eta)^2
\br{2-
\eta-2\psi(1-
\eta)}\\
&=
1+2\lambda
\eta\psi C^\star(
\eta)
+
\lambda^2
\eta C^\star(
\eta)^2
\br{
\eta\psi^2-
2+
\eta+2\psi(1-
\eta)}.
\end{align*}
The term in brackets equals \(-T_\eta\), and since \(C^\star(\eta)=\psi/T_\eta\), this becomes
\[
    1+2\lambda
\eta\psi C^\star(
\eta)
    -
    \lambda^2
\eta\psi C^\star(
\eta)
    =
    1+(2\lambda-
\lambda^2)
\eta\psi C^\star(
\eta).
\]
Combining this with \eqref{eq:rowMGFWithDp} completes the proof of \eqref{eq:rowMGFLimitSparsification}.
\end{proof}

\begin{proof}[Proof of Proposition \ref{proposition:sparsificationLargeDeviationBound}]
Fix \(S\in\calS\) with \(M(S)
\geq\delta s\), and let \(\eta=M(S)/s\). Apply Chernoff's bound \eqref{eq:ChernoffBoundSparsificationNew} with \(\theta=
\theta_{p,
\lambda}(
\eta)\). Lemma \ref{lemma:RowMGFLimitSparsification} gives, uniformly in \(S\),
\[
    \E\br{e^{-\theta_{p,
\lambda}(
\eta)
\Delta_1}}
    =
    \pr{1+(2\lambda-
\lambda^2)
\eta\psi C^\star(
\eta)}^{-1/2}+o(1).
\]
The limiting quantity is a constant in $\pr{0,1}$ for any \(\eta\in[
\delta,1]\), raising to the \(n\)-th power yields
\[
    \Pro\pr{
\Delta_S\leq0}
    \leq
    \pr{1+(2\lambda-
\lambda^2)
\eta\psi C^\star(
\eta)}^{-n/2}e^{o(n)},
\]
hence completing the proof of \eqref{eq:sparsificationLDlambda}.
\end{proof}

\section{Conclusion and Future Work}
\label{section:conclusion}
In the first part of this paper, we have studied the problem of recovery of a binary signal \(\beta^\star \in \ac{0,1}^p\) based on a sparse measurement matrix and noisy observations. Our main result is that, if the measurements have density rate \(d/p\) then, assuming that the measurements and the signal are together not too sparse -- in particular if \(ds = \omega\pr{p}\), i.e. high-SNR regime -- it is possible to recover the true support asymptotically when the sample size is larger than the threshold given by Theorem \ref{theorem:sparseRecoveryForSparseMeasurements}. Combining our work with the necessary conditions of Wang et al. \cite{wang2010information}, we reveal an information-theoretic phase transition. The expression of the phase-transition threshold makes the \emph{price of sparsity} explicit, revealing a precise trade‑off between sampling complexity and measurement sparsity. In the following, we present a quick summary of all -- to the best of our knowledge -- results on sparse recovery in the sparse measurement setting, along with some future work directions.

\begin{itemize}
\item Information-theoretic threshold, sufficient conditions. In Theorem \ref{theorem:sparseRecoveryForSparseMeasurements}, we establish a sufficient condition for reliable recovery. However, this result is conditional on the high-SNR ($ds/p \rightarrow +\infty$) assumption. This raises the question of sufficient conditions when the measurements and signal are even more sparse.
\item Informational-theoretic threshold, necessary conditions.
\begin{itemize}
\item Wang et al. \cite{wang2010information} have studied this problem. Their work reveals three regimes of behavior depending on the scaling of the expected number of non-zeros of \(\beta^\star\) aligning with non-zeros of a row of \(X\): \(ds/p = \omega\pr{1}\), \(ds/p \rightarrow \tau\) for some \(\tau > 0\), and \(ds/p = o\pr{1}\). For their model, where the variance of the non-zero components of \(X\) scales in a way that makes the second moment of the projected signal \(X \beta^\star\) remain the same as in the dense case: the necessary condition threshold is on the order of magnitude of the one in the dense case in the regime where \(ds/p = \omega\pr{1}\), while it increases dramatically in the regime where \(ds/p = o\pr{1}\).
\item In the dense setting, Reeves et al. \cite{reeves2019all} have shown that even the recovery of a fixed fraction of the support is information theoretically impossible below the phase transition threshold: this is what they call the \emph{all-or-nothing} property. It would be interesting to extend this property to the sparse setting.
\end{itemize}
\item Algorithmic threshold, sufficient conditions. Omidiran and Wainwright \cite{omidiran2008high} have shown that under a low-sparsity assumption on the measurements, the sufficient condition of the dense setting, i.e. \(n \geq \pr{1+\varepsilon}n_{\text{ALG}}\), is sufficient for the sparse setting as well. It would be interesting to explore the question of polynomial-time recovery in a stronger sparsity regime.
\item Algorithmic threshold, necessary conditions. Although we cannot really hope to provide necessary conditions for polynomial-time recovery -- unless conditionally on \(\text{P} \ne \text{NP}\) -- it would be interesting to provide a threshold under which the problem is believed to be algorithmically hard, as done by Gamarnik and Zadik \cite{gamarnik2022sparse} in the dense setting.
\end{itemize}

In the second part of this paper, we have studied the problem of recovering the signal based on \emph{sparsified} -- but originally dense -- measurements and accordingly-rescaled observations. Our main result is that, in the proportional sparsity and proportional sparsification regime where \(s = \alpha p\) and \(d = \psi p\), with \(\psi>0\) fixed and sufficiently small, it is possible to recover the true support asymptotically when the sample size is larger than a threshold given by Theorem \ref{theorem:sparsificationLinear}. This provides a sufficient recovery guarantee for all sufficiently small fixed \(\psi\), provided a large enough sample size, and gives an upper bound on the \emph{price of sparsification} in this regime.

Nevertheless, we believe that the sparsification problem is infeasible for strong enough regimes of sparsification. In particular, we conjecture that the recovery is information-theoretically impossible no matter the sample size in the sub-proportional sparsification regime where \(d = o\pr{p}\). We leave the exploration of this regime for future work.


\bibliographystyle{imsart-number} 
\bibliography{bibliography}       

\newpage

\appendix

\section{Omitted Proofs}

\subsection{Omitted Proofs from Theorem \ref{theorem:sparseRecoveryForSparseMeasurements}}

\subsubsection{Proof of Lemma \ref{lemma:AsymptoticBinomial}}
\label{section:proofOflemma:AsymptoticBinomial}
\begin{proof}[Proof of Lemma \ref{lemma:AsymptoticBinomial}]
Let:
\[
N_p \coloneqq
\frac{1}{\sqrt{2 \delta ds/p}}\pr{\sum_{j \in U_{[\delta s]} \cup V_{[\delta s]}} B_{ij} - 2 \delta ds/p}.
\]
Its characteristic function writes:
\begin{align*}
&\Phi_{N_p}\pr{t}\\
&= \E\br{e^{itN_p}}\\
&=
e^{-it\sqrt{2\delta ds/p}}
\pr{\E\br{\exp\pr{it B_{ij} / \sqrt{2\delta ds/p}}}}^{2\delta s}\\
&=
e^{-it\sqrt{2\delta ds/p}}
\pr{1 - d/p + d/p \exp\pr{it/\sqrt{2\delta ds/p}}}^{2\delta s}\\
&=
e^{-it\sqrt{2\delta ds/p}}
\pr{1 + d/p \pr{\exp\pr{it/\sqrt{2\delta ds/p}}-1}}^{2\delta s}\\
&=
e^{-it\sqrt{2\delta ds/p}}
\pr{1 + d/p \pr{\frac{it}{\sqrt{2\delta ds/p}}-\frac{t^2}{4\delta ds/p}+\calO\pr{\frac{-it^3}{\pr{2\delta ds/p}^{3/2}}}}}^{2\delta s}\\
&=
e^{-it\sqrt{2\delta ds/p}}
\pr{1 + d/p \pr{\frac{it}{\sqrt{2\delta ds/p}}-\frac{t^2}{4\delta ds/p}+\calO\pr{\frac{1}{\pr{2\delta ds/p}^{3/2}}}}}^{2\delta s}\\
&=
\exp\pr{
-it\sqrt{2\delta ds/p}+
2\delta s
\log\pr{1 + d/p \pr{\frac{it}{\sqrt{2\delta ds/p}}-\frac{t^2}{4\delta ds/p}+\calO\pr{\frac{1}{\pr{2\delta ds/p}^{3/2}}}}}
}\\
&=
e^{-it\sqrt{2\delta ds/p}}
\exp\pr{
2\delta s
\pr{d/p \pr{\frac{it}{\sqrt{2\delta ds/p}}-\frac{t^2}{4\delta ds/p}+\calO\pr{\frac{1}{\pr{2\delta ds/p}^{3/2}}}}
+\calO\pr{
\frac{d}{sp}}}}\\
&=
\exp\pr{
-it\sqrt{2\delta ds/p}+
it\sqrt{2\delta ds/p}
- t^2/2 +
\calO\pr{\frac{1}{\sqrt{2\delta ds/p}}}
+\calO\pr{
\frac{2\delta d}{p}}}\\
&=
\exp\pr{
- t^2/2 +
o\pr{1}}
\stackrel{p \rightarrow +\infty}{\longrightarrow} \exp\pr{
- t^2/2}
= \Phi_{\calN\pr{0,1}}\pr{t},
\end{align*}
for all \(t \in \R\). The result follows.
\end{proof}

\subsubsection{Proof of Lemma \ref{lemma:uniformIntegrabilityOfVp}}
\label{section:proofOflemma:uniformIntegrabilityOfVp}
\begin{proof}[Proof of Lemma \ref{lemma:uniformIntegrabilityOfVp}]
Write \(\Sigma_p \coloneqq \sum_{j \in U_{[\delta s]} \cup V_{[\delta s]}} B_{ij}\sim\text{Bin}\pr{2\delta s, d/p}\),
with mean \(\mu_p \coloneqq \E\br{\Sigma_p} = 2\delta ds/p\). We recall:
\[
N_p = \frac{\Sigma_p - \mu_p}{\sqrt{\mu_p}},
\qquad
V_p = \frac{2\sigma}{\sqrt{2\delta + \frac{4\sigma^2}{ds/p} + \sqrt{\frac{2\delta}{ds/p}}\,N_p}},
\]
and note the deterministic bound \(0 \leq V_p \leq \sqrt{ds/p}\), which holds because \(\Sigma_p \geq 0\). Fix \(T > 2\). Splitting on
\(\calE_p \coloneqq \acn{N_p \geq -\pr{ds/p}^{1/4}}\), we have:
\[
\E\br{V_p\,\indic\ac{V_p > T}} = \Xi_1 + \Xi_2,
\quad
\Xi_1 \coloneqq \E\br{V_p\,\indic\ac{V_p > T}\,\indic_{\calE_p}},
\quad
\Xi_2 \coloneqq \E\br{V_p\,\indic\ac{V_p > T}\,\indic_{\calE_p^c}}.
\]

\emph{Bounding \(\Xi_1\).} On \(\calE_p\) the denominator defining \(V_p\) satisfies
\[
2\delta + \frac{4\sigma^2}{ds/p} + \sqrt{\frac{2\delta}{ds/p}}\,N_p
\;\geq\;
2\delta + \frac{4\sigma^2}{ds/p} - \sqrt{2\delta}\,\pr{\frac{ds}{p}}^{-1/4}.
\]
The right-hand side is deterministic and converges to \(2\delta\), so there is \(p_1 \in \N\) with
right-hand side \(\geq \delta\) for all \(p \geq p_1\). Hence \(V_p \leq 2\sigma/\sqrt{\delta}\) on
\(\calE_p\) for \(p \geq p_1\), so \(\indic\ac{V_p > T}\indic_{\calE_p} \leq \indic\ac{2\sigma/\sqrt{\delta} > T}\) and
\[
\Xi_1 \leq \frac{2\sigma}{\sqrt{\delta}}\,\indic\ac{2\sigma/\sqrt{\delta} > T},
\qquad p \geq p_1.
\]

\emph{Bounding \(\Xi_2\).} By \(V_p \leq \sqrt{ds/p}\) we have \(\indic\ac{V_p > T} \leq \indic\ac{ds/p > T^2}\), and therefore:
\[
\Xi_2 \leq \sqrt{ds/p}\,\indic\ac{ds/p > T^2}\,\Pro\pr{\calE_p^c}.
\]
The complementary event is a Binomial lower-tail deviation: \(\calE_p^c = \acn{\Sigma_p < \pr{1-\varepsilon_p}\mu_p}\)
with \(\varepsilon_p \coloneqq \pr{ds/p}^{-1/4}/\sqrt{2\delta} \in (0,1)\) for \(p\) large. The multiplicative
Chernoff bound \(\Pro\pr{\Sigma_p \leq \pr{1-\varepsilon}\mu_p} \leq e^{-\varepsilon^2\mu_p/2}\) gives
\[
\Pro\pr{\calE_p^c} \leq \exp\pr{-\frac{\varepsilon_p^2 \mu_p}{2}} = \exp\pr{-\frac{\sqrt{ds/p}}{2}},
\qquad
\varepsilon_p^2 \mu_p = \frac{1}{2\delta\sqrt{ds/p}}\cdot 2\delta\,\frac{ds}{p} = \sqrt{ds/p}.
\]
Hence, for \(p \geq p_2\) (large enough that \(\varepsilon_p \in (0,1)\)), we have:
\[
\Xi_2 \leq \sqrt{ds/p}\,e^{-\sqrt{ds/p}/2}\,\indic\ac{ds/p > T^2}.
\]
On \(\acn{ds/p > T^2}\) we have
\(\sqrt{ds/p} > T > 2\), and \(x \mapsto \sqrt{x}\,e^{-\sqrt{x}/2}\) is decreasing for \(\sqrt{x} > 2\), so
\(\Xi_2 \leq T e^{-T/2}\).\\

We conclude that, for all \(p \geq p' \coloneqq \max\pr{p_1, p_2}\), we have:
\[
\E\br{V_p\,\indic\ac{V_p > T}} \leq \frac{2\sigma}{\sqrt{\delta}}\,\indic\ac{2\sigma/\sqrt{\delta} > T} + T e^{-T/2}.
\]
The right-hand side is independent of \(p\) and tends to \(0\) as \(T \rightarrow +\infty\) (the first term
vanishes once \(T > 2\sigma/\sqrt{\delta}\)). Therefore
\[
\lim_{T \rightarrow +\infty}\,\sup_{p \geq p'}\,\E\br{V_p\,\indic\ac{V_p > T}} = 0.
\]
\end{proof}


\subsection{Omitted Proofs from Theorem \ref{theorem:sparsificationLinear}}

\subsubsection{Derivation of the conditional row moment generating function}
\label{section:conditionalChernoffExpansion}

\begin{proof}[Derivation of \eqref{eq:conditionalMGFPreUV}]
We work with the first row and abbreviate \(X_j\equiv X_{1j}\),
\(B_j\equiv B_{1j}\), \(Z\equiv Z_1\). Since \(\indic_{S^\star}=\indic_A+\indic_C\),
\(\indic_S=\indic_B+\indic_C\), and \(\Tilde X_{1j}=B_jX_j\), set
\[
W\coloneqq \sum_{j\in A}B_jX_j-\sum_{j\in B}B_jX_j
=\inner{\Tilde X_1}{\indic_{S^\star}-\indic_S},
\]
\[
R\coloneqq \sum_{j\in A}B_jX_j+\sum_{j\in B}B_jX_j+2\sum_{j\in C}B_jX_j
=\inner{\Tilde X_1}{\indic_{S^\star}}+\inner{\Tilde X_1}{\indic_S},
\]
and
\[
D\coloneqq \sum_{j\in S^\star}X_j=\inner{X_1}{\indic_{S^\star}}.
\]

\emph{Expanding \(\Delta_1\).} Factoring the difference of squares in the
definition of \(\Delta_1\) and using
\(\inner{\Tilde X_1}{\indic_S}-\inner{\Tilde X_1}{\indic_{S^\star}}=-W\), we get:
\[
\inner{\Tilde X_1}{\indic_S}^2-\inner{\Tilde X_1}{\indic_{S^\star}}^2
=\pr{\inner{\Tilde X_1}{\indic_S}-\inner{\Tilde X_1}{\indic_{S^\star}}}
 \pr{\inner{\Tilde X_1}{\indic_S}+\inner{\Tilde X_1}{\indic_{S^\star}}}
=-WR,
\]
while the cross term equals \(2\psi\pr{D+Z}W\). Hence
\[
\Delta_1
=-WR+2\psi\pr{D+Z}W
=W\pr{2\psi D-R}+2\psi W Z,
\]
and therefore:
\[
-\theta\Delta_1=\theta W\pr{R-2\psi D}-2\theta\psi W Z.
\]

\emph{Integrating out \(Z\).} Only the final term involves \(Z\). Conditioning on
\((X_1,B_1)\) and integrating \(Z\sim\calN(0,\sigma^2)\),
\[
\E\br{e^{-2\theta\psi W Z}\mid X_1,B_1}
=e^{\frac12\pr{2\theta\psi W}^2\sigma^2}
=e^{2\theta^2\psi^2\sigma^2W^2}
=e^{\theta\gamma(\theta)W^2},
\]
where the last equality holds by \(\gamma(\theta)=2\psi^2\sigma^2\theta\) from
\eqref{equation:definitionGamma}. Therefore
\begin{align*}
\E\br{e^{-\theta\Delta_1}\mid B_1}
&=\E_{X_1}\br{\exp\pr{\theta W\pr{R-2\psi D}+\theta\gamma(\theta)W^2}\mid B_1}\\
&=\E_{X_1}\br{\exp\pr{\theta W\pr{R-2\psi D+\gamma(\theta)W}}\mid B_1}.
\end{align*}
With \(U_\theta\coloneqq\theta W\) and
\(V_\theta\coloneqq R-2\psi D+\gamma(\theta)W\), the exponent is \(U_\theta V_\theta\),
which is \eqref{eq:conditionalMGFPreUV}.\\

\emph{Explicit form of \(U_\theta,V_\theta\).} Now we have:
\[
U_\theta=\theta W=\theta\pr{\sum_{j\in A}B_jX_j-\sum_{j\in B}B_jX_j}.
\]
Regrouping
\(V_\theta\) by index set, we obtain:
\begin{align*}
V_\theta
&=\pr{\sum_{j\in A}B_jX_j+\sum_{j\in B}B_jX_j+2\sum_{j\in C}B_jX_j}
-2\psi\pr{\sum_{j\in A}X_j+\sum_{j\in C}X_j}\\
&\qquad +\gamma(\theta)\pr{\sum_{j\in A}B_jX_j-\sum_{j\in B}B_jX_j}\\
&=\sum_{j\in A}\pr{\pr{1+\gamma(\theta)}B_j-2\psi}X_j
+\sum_{j\in B}\pr{1-\gamma(\theta)}B_j X_j
+2\sum_{j\in C}\pr{B_j-\psi}X_j,
\end{align*}
matching the definitions in
Section \ref{section:proofOftheorem:sparsificationLinear}. Since \(U_\theta\) and
\(V_\theta\) are linear in the Gaussian vector \(X_1\) with coefficients
determined by \(B_1\), the pair \((U_\theta,V_\theta)\) is, conditionally on
\(B_1\), centered and jointly Gaussian.
\end{proof}

\subsubsection{Conditional variances and covariance of the Gaussian pair}
\label{section:conditionalMomentsUV}

\begin{proof}[Derivation of \eqref{eq:apSparsification}, \eqref{eq:bpSparsification} and \eqref{eq:cpSparsification}]
Fix a mask realization \(B_1=b\) and abbreviate \(X_j\equiv X_{1j}\),
\(\gamma\equiv\gamma(\theta)\). Conditionally on \(B_1=b\), the \(X_j\) are
independent \(\calN(0,1)\), so \(U_\theta\) and \(V_\theta\), being linear in
\((X_j)_{j\in[p]}\), are centered and jointly Gaussian, and their conditional second moments are obtained by summing products of their \(X_j\)-coefficients over
\(j\), the sets \(A,B,C\) being disjoint. Throughout we use \(b_j\in\{0,1\}\), hence
\(b_j^2=b_j\), together with \(\abs{A}+\abs{C}=\abs{S^\star}=s\). Reading off the
definitions of \(U_\theta\) and \(V_\theta\) in
Section~\ref{section:proofOftheorem:sparsificationLinear}, the \(X_j\)-coefficients
of \(U_\theta\) are \(\theta b_j\) on \(A\), \(-\theta b_j\) on \(B\), and \(0\) on
\(C\); those of \(V_\theta\) are \((1+\gamma)b_j-2\psi\) on \(A\), \((1-\gamma)b_j\)
on \(B\), and \(2(b_j-\psi)\) on \(C\).\\

\emph{Variance of \(U_\theta\).} Summing the squared coefficients, we obtain:
\[
\var(U_\theta\mid B_1=b)
=\theta^2\sum_{j\in A}b_j^2+\theta^2\sum_{j\in B}b_j^2
=\theta^2\pr{x_A(b)+x_B(b)},
\]
which is \eqref{eq:apSparsification}.\\

\emph{Variance of \(V_\theta\).} The per-coordinate squared coefficients are
\[
\begin{aligned}
j\in A:&\quad \pr{(1+\gamma)b_j-2\psi}^2
=\pr{(1+\gamma)^2-4\psi(1+\gamma)}b_j+4\psi^2,\\
j\in B:&\quad \pr{(1-\gamma)b_j}^2=(1-\gamma)^2 b_j,\\
j\in C:&\quad \pr{2(b_j-\psi)}^2=4(1-2\psi)b_j+4\psi^2,
\end{aligned}
\]
using \(b_j^2=b_j\). Summing over each set and combining the constant \(4\psi^2\) terms through \(\abs{A}+\abs{C}=s\), we obtain:
\[
\var(V_\theta\mid B_1=b)
=4\psi^2 s
+\pr{(1+\gamma)^2-4\psi(1+\gamma)}x_A(b)
+(1-\gamma)^2 x_B(b)
+4(1-2\psi)x_C(b),
\]
which is \eqref{eq:bpSparsification}.\\

\emph{Covariance.} The \(U_\theta\)-coefficient vanishes on \(C\), so only \(A\) and
\(B\) contribute:
\begin{align*}
\cov(U_\theta,V_\theta\mid B_1=b)
&=\sum_{j\in A}(\theta b_j)\pr{(1+\gamma)b_j-2\psi}
+\sum_{j\in B}(-\theta b_j)\pr{(1-\gamma)b_j}\\
&=\theta\sum_{j\in A}\pr{(1+\gamma)-2\psi}b_j-\theta\sum_{j\in B}(1-\gamma)b_j\\
&=\theta\br{\pr{1+\gamma-2\psi}x_A(b)-\pr{1-\gamma}x_B(b)},
\end{align*}
which is \eqref{eq:cpSparsification}.
\end{proof}

\subsubsection{Proof of Lemma \ref{lemma:GaussianProductMGFNew}}
\label{section:proofOflemma:GaussianProductMGFNew}

\begin{proof}[Proof of Lemma \ref{lemma:GaussianProductMGFNew}]
The covariance matrix \(\Sigma=\begin{pmatrix}a&c\\c&b\end{pmatrix}\) is positive semidefinite, so \(\det\Sigma=ab-c^2\ge 0\). We split according to whether \(\Sigma\) is positive definite (\(ab>c^2\)) or degenerate (\(ab=c^2\)); the degenerate case is further divided into two sub-cases.\\

\emph{Case 1: \(\Sigma\succ 0\).} For \(G=(U,V)^\T\), since \(UV=\frac12G^\T JG\), where
\[
    J=\begin{pmatrix}0&1\\1&0\end{pmatrix},
\]
the standard quadratic-form Gaussian MGF formula gives
\[
    \E\br{e^{UV}}
    =
    \det(I-\Sigma J)^{-1/2}
\]
provided \(\Sigma^{-1}-J\succ 0\); otherwise the integral diverges. Since \(\det(\Sigma^{-1}-J)=\det(\Sigma^{-1})\det(I-\Sigma J)=((1-c)^2-ab)/\det(\Sigma)\) and \(\det(\Sigma)=ab-c^2>0\) by positive definiteness, the condition \(\Sigma^{-1}-J\succ 0\) is equivalent to \(D=(1-c)^2-ab>0\). Computing
\[
    I-\Sigma J
    =
    \begin{pmatrix}1-c&-a\\-b&1-c\end{pmatrix},
\]
we obtain \(\det(I-\Sigma J)=(1-c)^2-ab=D\), proving the claim.\\

\emph{Case 2: \(\Sigma\) is degenerate with \(a=0\) or \(b=0\).} Then \(c^2 = ab\) forces \(c=0\), so \(U=0\) a.s.\ or \(V=0\) a.s. In either case, \(UV=0\) a.s.\ and \(\E[e^{UV}]=1\), which agrees with \(D=(1-c)^2-ab=1\), so \(D^{-1/2}=1\).\\

\emph{Case 3: \(\Sigma\) is degenerate with \(a,b>0\).} Then \(c^2=ab\), so \(V=(c/a) \, U\) a.s.\ with \( U \sim \calN(0,a)\). Hence \(UV=(c/a)U^2\) where \(U^2/a\sim\chi^2_1\), giving
\[
    \E\br{e^{UV}}
    =\E\br{e^{c\cdot U^2/a}}
    =(1-2c)^{-1/2}\quad\text{if }c<1/2,
\]
and \(+\infty\) otherwise. On the other hand,
\[
    D
    =(1-c)^2-ab
    =(1-c)^2-c^2
    =1-2c,
\]
so the formula \(\E[e^{UV}]=D^{-1/2}\) holds, and the \(D > 0\) condition matches.
\end{proof}

\end{document}